\pdfoutput=1
\documentclass[11pt]{article}

\usepackage{EMNLP2023}

\usepackage{times}
\usepackage{latexsym}

\usepackage[T1]{fontenc}
\usepackage[utf8]{inputenc}

\usepackage{microtype}

\usepackage{algorithm}
\usepackage{algorithmic}
\usepackage{booktabs}
\usepackage{multirow}
\usepackage{graphicx}
\usepackage{subfigure}
\usepackage{caption}
\usepackage{amsmath}

\usepackage{inconsolata}

\title{Towards Informative Few-Shot Prompt with Maximum Information Gain for In-Context Learning}

\author{Hongfu Liu \and Ye Wang \\
        School of Computing, National University of Singapore \\ \texttt{\{hongfu,wangye\}@comp.nus.edu.sg}}
\begin{document}
\maketitle

\begin{abstract}
% Large Language models (LLMs) are capable of performing In-context Learning (ICL) with a few demonstrations of a new downstream task as conditions. However, such a learning paradigm suffers from high variances caused by the distribution of selected examples, their ordering, and prompt formats. We show that even if all of them are fixed, randomly selecting examples still incurs substantial variance. In this work, we investigate the informative ability of data examples by measuring the information gain of prediction after observing the example candidate. Then we propose to sample those with maximum information gain. Besides, we find that the template bias can lead to unfair evaluations of information gain when sampling and propose calibration before sampling to address it. Our method achieves 14.3\% relative improvements on average across six classification tasks and three LLMs.
Large Language models (LLMs) possess the capability to engage In-context Learning (ICL) by leveraging a few demonstrations pertaining to a new downstream task as conditions. However, this particular learning paradigm suffers from high instability stemming from substantial variances induced by factors such as the input distribution of selected examples, their ordering, and prompt formats. In this work, we demonstrate that even when all these factors are held constant, the random selection of examples still results in high variance. Consequently, we aim to explore the informative ability of data examples by quantifying the Information Gain (IG) obtained in prediction after observing a given example candidate. Then we propose to sample those with maximum IG. Additionally, we identify the presence of template bias, which can lead to unfair evaluations of IG during the sampling process. To mitigate this bias, we introduce Calibration Before Sampling strategy. The experimental results illustrate that our proposed method can yield an average relative improvement of 14.3\% across six classification tasks using three LLMs.
\end{abstract}

\section{Introduction}

% background - LLM ICL
% These days the In-context Learning (ICL) ability of pre-trained Large Language Models (LLMs) has received much attention in the community. ICL is a new paradigm for few-shot learning, referring to the process of performing new downstream tasks conditioned on prompts. Such prompts are comprised of a few input-output pairs from annotated training examples, commonly referred to as demonstrations, and serve as a task description for the LLM. LLMs have showcased the formidable capacity of ICL and achieved remarkable performances across various downstream tasks~\cite{brown2020language}. In comparison to approaches that involve fine-tuning LLMs on downstream tasks~\cite{devlin-etal-2019-bert,gao-etal-2021-making}, ICL requires no parameter updates, allowing higher computational efficiency and easier deployment. 
These days the In-context Learning (ICL) ability of pre-trained Large Language Models (LLMs) has garnered significant attention in the community. ICL represents a new paradigm for few-shot learning, which entails performing new downstream tasks based on prompts. These prompts consist of a few input-output pairs, commonly referred to as demonstrations. Such prompts serve as explicit task descriptions for the LLM. LLMs have showcased the formidable capacity of ICL and achieved remarkable performances across various downstream tasks~\cite{brown2020language}. In comparison to approaches that involve fine-tuning LLMs on downstream tasks~\cite{devlin-etal-2019-bert,gao-etal-2021-making}, ICL obviates the need for parameter updates, thereby allowing higher efficiency in adapting to new tasks and easier deployment. 

% \begin{figure}[t]
%   \centering
%   \subfigure[]{
%       \includegraphics[scale=0.36]{emnlp2023-latex/img/var_gpt2.pdf}
%   }\hspace{-3mm}
%   \subfigure[]{
%       \includegraphics[scale=0.36]{emnlp2023-latex/img/var_gptj.pdf}
%   }
%   \caption{Four-shot ICL performance of GPT-2 XL and GPT-J on SST-2. Each boxplot summarizes the results of 10 randomly sampled prompts for one sample order. Given one sample order, the input distribution, demonstration ordering, and prompt formats are fixed. For example, one sample order [P P N N] denotes two positive examples and then two negative ones. }
%   \label{fig:var}
%   \vspace{-1mm}
% \end{figure}
\begin{figure}[t]
  \centering
  \subfigure{
      \includegraphics[scale=0.36]{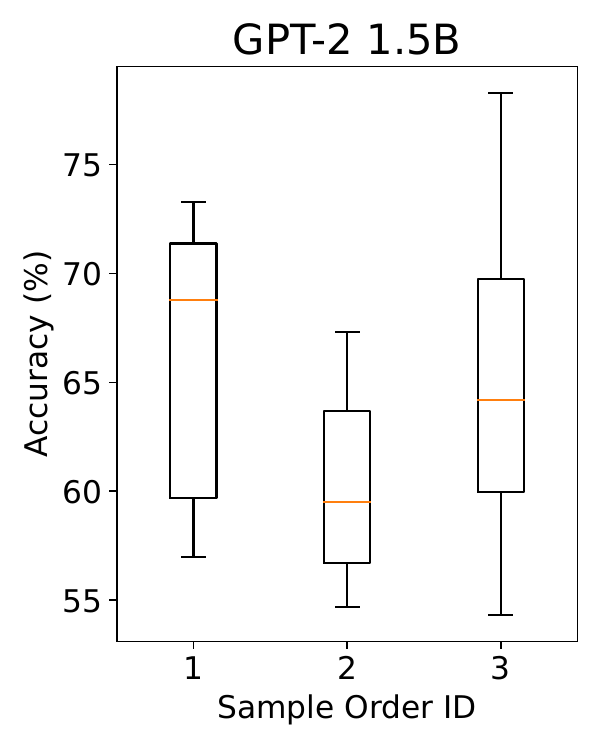}
  }\hspace{-3mm}
  \subfigure{
      \includegraphics[scale=0.36]{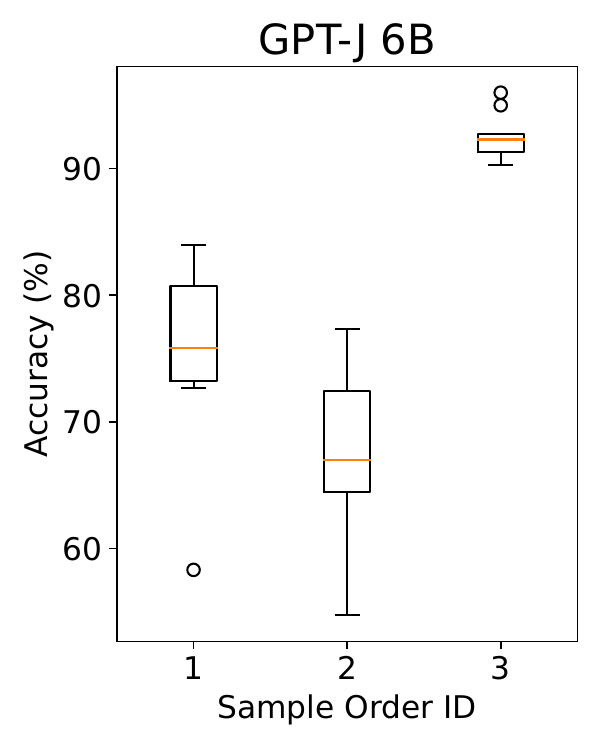}
  }
  \caption{Four-shot ICL performance on SST-2 using GPT-2 XL and GPT-J. Each boxplot summarizes the results of 10 randomly selected prompts for a specific sample order. For a given sample order, the input distribution, demonstration ordering, and prompt formats remain constant. For instance, the sample order [P P N N] denotes the sequence of two positive examples followed by two negative ones. }
  \label{fig:var}
  % \vspace{-2mm}
\end{figure}

% challenge
% However, ICL tends to suffer from substantial variance in performance. Existing studies attribute it to the input distribution of demonstrations, their ordering, and the prompt formats when constructing prompts~\cite{zhao2021calibrate,lu-etal-2022-fantastically,zhang-etal-2022-active,min-etal-2022-rethinking}. We find that even if the input distribution, the ordering of demonstrations, and the prompt formats are fixed, randomly selecting different demonstrations still causes high variance as is shown in Figure \ref{fig:var}. This indicates that data samples within the same category can provide different information and contribute differently to the ICL performance. We refer to the information provided by one data sample as the informative ability. As far as we know, there is no prior work investigating it in the literature. 
However, ICL tends to suffer from substantial variance in performance. Existing studies attribute it to factors including the input distribution of demonstrations, their ordering, and the prompt formats employed during prompt construction~\cite{zhao2021calibrate,lu-etal-2022-fantastically,zhang-etal-2022-active,min-etal-2022-rethinking}. Our investigation reveals that even when the input distribution, the ordering of demonstrations, and prompt formats remain fixed, the random selection of different demonstrations still leads to significant variance, as shown in Figure \ref{fig:var}. This observation indicates that data samples within the same category can offer distinct information and contribute differently to the ICL performance. We refer to the ability of a data sample to provide valuable information as its informative ability. To the best of our knowledge, there has been no prior study exploring this aspect in the existing literature. 

% This indicates that the informative ability of data within the same category is different and the difference between them is crucial for ICL as well. As far as we know, there is no prior work investigating it in the literature. 

% proposed method
% In this paper, we examine the informative ability of data examples from the aspect of information theory and investigate the association between this ability with the ICL performance. Specifically, we assess the extent to which one data example contributes information to the specific downstream task. To measure the information ability of one example candidate quantitatively, we propose to evaluate the Information Gain (IG) of prediction, which quantifies how much information is obtained after observing one example candidate in the context. To achieve this, we construct a prompt for each example candidate and leverage the LLM to obtain the corresponding output prediction and evaluate IG. Furthermore, we discover that Template Bias may lead to unfair evaluations of IG. To rectify this, we introduce a Calibration Before Sampling strategy to ensure a fair assessment of IG. Subsequently, we select the example candidates with maximum IG and annotate them as demonstrations for ICL. 
In this paper, we examine the informative ability of data examples from the perspective of information theory and investigate its correlation with the ICL performance. Specifically, we assess the contribution of individual data samples to the specific downstream tasks by quantitatively measuring their informative ability. To accomplish this, we propose to evaluate the Information Gain (IG) of prediction, which quantifies the amount of information gained after observing one example candidate in the context. We construct a prompt for each example candidate and utilize the LLM to obtain the corresponding output distribution, thus enabling the evaluation of IG. Furthermore, we uncover the presence of Template Bias, which can lead to biased evaluations of IG. To address this issue, we introduce a Calibration Before Sampling strategy to ensure a fair assessment of IG. Subsequently, we select the example candidates with maximum IG and annotate them as demonstrations for enhancing ICL. 

% We empirically evaluate the effectiveness of our method across six classification datasets and three LLMs with various model sizes. Experimental results demonstrate that we can achieve 14.3\% relative improvement on average. Importantly, it should be noted that our proposed method is orthogonal to existing calibration~\cite{zhao2021calibrate} and reordering methods~\cite{lu-etal-2022-fantastically}, and we demonstrate that our method can be combined with these approaches for further improvements. 

To validate the effectiveness of our method, we conduct empirical evaluations across six classification datasets across three LLMs of varying model sizes. The experimental results demonstrate an average relative improvement of 14.3\% on one-shot learning. It is important to emphasize that our proposed method is orthogonal to existing methods such as calibration~\cite{zhao2021calibrate} and reordering methods~\cite{lu-etal-2022-fantastically}. Moreover, we demonstrate that our method can be combined with these approaches to achieve further improvements. Additionally, we analyze the relationship between data informative ability with the correctness of target labels and find that data examples with high IG tend to rely more on the accuracy of target labels.

% Overall, our contribution can be summarized as follows:
% \begin{itemize}
%     \item We study the association between data informative ability and In-context Learning performance.
%     \item We propose to measure the data informative ability based on Information Gain (IG) and sample the demonstrations with maximum IG to improve ICL performance.
%     \item We recognize the Template Bias and propose Calibration Before Sampling to rectify it.
%     \item Our proposed method is effective across six classification tasks across three LLMs with 14.3\% relative improvement on average.  
% \end{itemize}

In summary, our contributions can be summarized as follows:
\begin{itemize}
    \item We investigate the relationship between data informative ability and ICL performance.
    \item We propose the use of Information Gain (IG) to measure the data informative ability and select demonstrations with maximum IG to enhance ICL performance.
    \item We identify Template Bias and introduce the Calibration Before Sampling strategy to address it.
    \item Our proposed method yields significant improvements, achieving an average relative improvement of 14.3\% across six classification tasks using three LLMs.  
\end{itemize}

\section{Related Work}

\subsection{Active Data Sampling for ICL}
% \textbf{Active Data Sampling for ICL} 
Active data sampling has been employed in natural language processing tasks since their early stage~\cite{settles2009active}. The primary objective is to achieve comparable or superior performance while reducing the annotation cost. With the advent of pre-trained LLMs, recent studies~\cite{alemp,cold,contrastive,yu2022cold} have successfully introduced active learning to minimize the amount of data required for fine-tuning. In the context of the ICL paradigm, the standard ICL involves the random selection of training examples as prompts. However, it has been observed that the performance and stability of ICL can be enhanced by selecting high-quality examples, which aligns with the concept of active data sampling. One common method is to retrieve semantically similar samples for each test query~\cite{rubin2021learning,liu2021makes,hongjin2023selective}, thereby enabling the utilization of instance-level prompts in downstream tasks. Another approach involves retrieving task-level examples as prompts for all test samples, eliminating the need for instance-level retrieval. \cite{zhang-etal-2022-active} introduces reinforcement learning to learn a generalized policy for example selection and \cite{chang2022careful} focuses on carefully choosing training subsets to improve stability. However, previous studies either rely on the performances of validation sets as reward signals or require the training of additional models to score example candidates. In contrast, our work centers on retrieving task-level examples from unlabeled datasets for all test samples, without the necessity of extra validation sets or the training of additional models.

\subsection{Confidence-based ICL Selection} 
% \textbf{Confidence-based ICL Selection} 
Confidence-based evaluation is widely used for in-context examples. Prior works take the confident model outputs as in-context examples~\cite{wanaclfinding} and search confident in-context example organizations in a self-adaptive manner~\cite{wu-etal-2023-self}. In concurrent works, USP~\cite{wan2023universal} utilizes confidence-based prediction for pseudo-demos generation and also handles generation tasks. LENS~\cite{li2023finding} proposed informative examples filtering and diversity-guided search method. Our work focuses on addressing the template bias of ICL selection by calibration before sampling.

\begin{figure*}[htbp]
  \centering
  \includegraphics[scale=0.45]{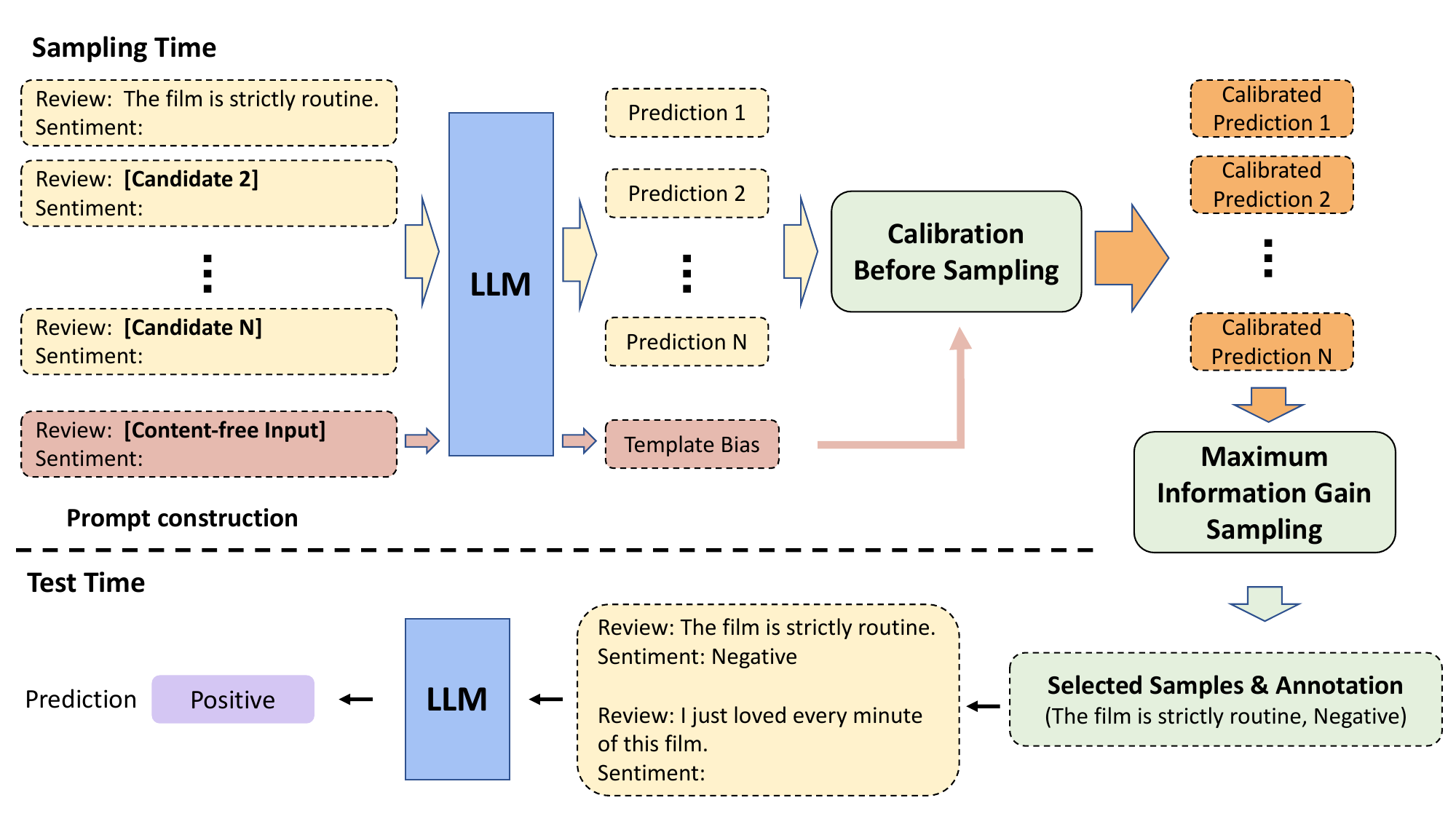}
  \caption{An overview of our proposed method. In Sampling Time, we construct zero-shot prompts, make predictions of example candidates, estimate Template Bias, perform calibration, and sample examples with maximum IG. In Test Time, we annotate selected samples and perform few-shot learning for test samples. \textbf{[Candidate N]} denotes the N-th example in the $\mathcal{D}_{unlab}$. \textbf{[Content-free Input]} denotes the content-free strings in $\mathcal{D}_{cf}$.}
  \label{fig:main_fig}
  \vspace{-2mm}
\end{figure*}

\section{Methodology}
\subsection{Problem Statement}
% We consider the setup that is close to true few-shot learning~\cite{perez2021true}. In order to retrieve prompts from the unlabeled text dataset $\mathcal{D}_{unlab} =\{x_i\}_{i=1}^N$ for a certain task, we use the pre-trained LLM to make inferences on all example candidates in $\mathcal{D}_{unlab}$ to get the corresponding prediction $\mathcal{Y}=\{\mathbf{y}_i\}_{i=1}^N$ where $\mathbf{y}_i$ is the output distribution given $x_i$. Our goal is to retrieve a subset $\{x_j\}_{j=1}^K$ of $\mathcal{D}_{unlab}$ where $K \ll N$ to facilitate few-shot learning, specifically, $K$-shot. We annotate the selected $K$ examples with their corresponding target labels $y^t$ and use these annotated examples $\mathcal{D}_{lab} = \{x_j, y^t_j\}_{j=1}^K$ as task-level prompts constructed using task-specific formats (See Appendix for details on formats we used), which appears as the prefix sequence for test samples. 
In this study, we focus on a problem setting that closely resembles true few-shot learning~\cite{perez2021true}, which is to retrieve prompts from an unlabeled text dataset, denoted as $\mathcal{D}_{unlab} =\{x_i\}_{i=1}^N$, for a specific task. To accomplish it, we utilize a pre-trained LLM to make predictions on all candidate examples in $\mathcal{D}_{unlab}$, resulting in the corresponding prediction set $\mathcal{Y}=\{\mathbf{y}_i\}_{i=1}^N$, where $\mathbf{y}_i$ represents the normalized predicted label distribution given input $x_i$. The goal is to select a subset $\{x_j\}_{j=1}^K$ from $\mathcal{D}_{unlab}$, where $K \ll N$, in order to facilitate few-shot learning, specifically, $K$-shot learning. We annotate the chosen $K$ examples with their respective target labels $y^t$ and construct task-level prompts using the input-label pairs and task-specific formats (See Appendix \ref{app_template} for details). The task-specific prompts are then incorporated as the prefix sequences for test samples. 

% $\mathcal{D}_{lab} = \{x_j, y^t_j\}_{j=1}^K$
\subsection{Information Gain}
% Information Gain (IG) is a measure of the amount of information gained about a random variable from observing another random variable~\cite{ash2012information}. To measure the informative ability of the data example, here we define the IG as the information obtained in prediction distribution $Y$ when observing one example candidate $X=x_{ob}$ in $\mathcal{D}_{unlab}$. Specifically, 
% \begin{equation}
%     IG(Y, x_{ob}) = H(Y) - H(Y|x_{ob})
% \end{equation}
% where $H(Y)$ is the information entropy of $Y$ and $H(Y|x_{ob})$ denotes the conditional entropy of $Y$ given the observation $x_{ob}$. However, the exact value of $IG(Y, x_{ob})$ is intractable due to the unknown $H(Y)$. Fortunately, $H(Y)$ is a constant value given a specific task. As a result, sampling examples with maximum information gain is equivalent to sampling those with minimum conditional entropy $H(Y|x_{ob})$. Specifically, given the LLM parameterized by $\theta$, 
% \begin{equation}
%     H(Y|x_{ob}) = - \sum_{y\in Y} p_{\theta}(y|x_{ob})\log p_{\theta}(y|x_{ob})
% \end{equation}
% However, it is difficult to compute $p_{\theta}(y|x_{ob})$ directly by inputting $x_{ob}$ into the LLM. Instead, we construct the prompt in the way of performing zero-shot ICL. One example is shown in the prompt construction of Figure \ref{fig:main_fig}. We fill in the template with text input only and let the LLM make predictions. In other words, we take each example candidate as the test sample in zero-shot ICL. As such, $p_{\theta}(y|x_{ob})$ is actually approximated by $p_{\theta}(y|x_{ob}, T)$ where $T$ is the task template.
Information Gain (IG) serves as a metric to quantify the amount of information obtained about a random variable through the observation of another random variable~\cite{ash2012information}. In our context, to measure the informative ability of data examples, we define the IG as the information obtained in predicted label distribution $Y$ when observing one example candidate $X=x_{ob}$ in $\mathcal{D}_{unlab}$. Specifically, 
\begin{equation}
    IG(Y, x_{ob}) = H(Y) - H(Y|x_{ob})
\end{equation}
where $H(Y)$ represents the information entropy of $Y$ and $H(Y|x_{ob})$ denotes the conditional entropy of $Y$ given the observation $x_{ob}$. However, computing the exact value of $IG(Y, x_{ob})$ is intractable due to the unknown $H(Y)$. Fortunately, $H(Y)$ remains constant for a given task, allowing us to reframe the problem of sampling examples with maximum IG as selecting those with minimum conditional entropy $H(Y|x_{ob})$. Specifically, considering the LLM parameterized by $\theta$, 
\begin{equation}
    H(Y|x_{ob}) = - \sum_{y\in Y} p_{\theta}(y|x_{ob})\log p_{\theta}(y|x_{ob})
\end{equation}
However, it is challenging to compute $p_{\theta}(y|x_{ob})$ directly by inputting $x_{ob}$ into the LLM. Instead, we adopt the approach of constructing the zero-shot prompt. One example is shown in the prompt construction of Figure \ref{fig:main_fig}. We fill in the task template with text input only and utilize the LLM to make predictions. In other words, each example candidate is taken as a test sample in zero-shot ICL. As such, $p_{\theta}(y|x_{ob})$ is actually approximated by $p_{\theta}(y|x_{ob}, T)$, where $T$ denotes the task template.

% check template bias
\subsection{Template Bias}
\begin{figure}[htbp]
  \centering
  \includegraphics[scale=0.3]{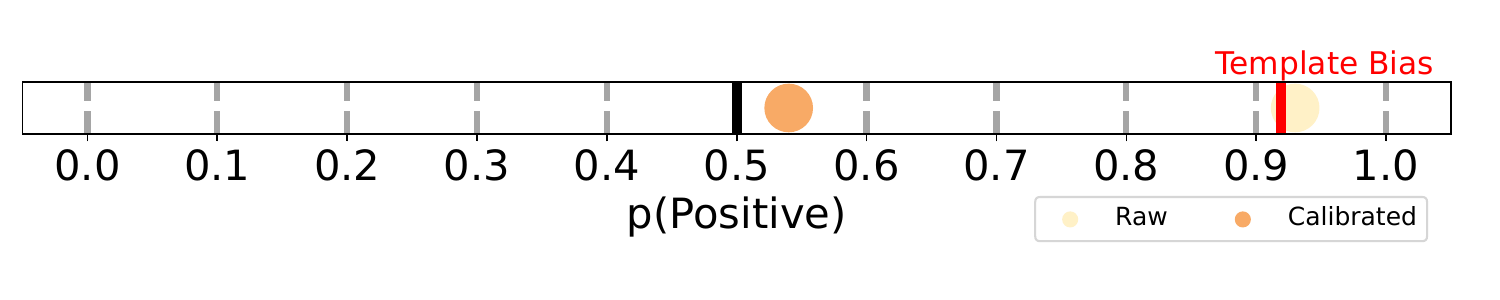}
  \caption{Template Bias on SST-2. The "Raw" point refers to the one before calibration. The "Calibrated" point represents the one after calibration. The balanced line (p=0.5) is bolded.}
  \label{fig:template_bias}
\end{figure}

% Inspired by \cite{zhao2021calibrate}, we find the predictive bias also exists when only inputting the template into the LLM. The red line in Figure \ref{fig:template_bias} demonstrates the bias of the template we used in SST-2 where the possibility of being positive is over 90\% given context-free input (discussed in Section \ref{cbs} ). We call it Template Bias \footnote{We use the term "Template Bias" to distinguish it from the Bias mentioned in \cite{zhao2021calibrate}.}, which refers to the zero-shot prompt bias. This indicates that even without any demonstration, the template used to prompt models to perform a certain task, also makes the model do predictions toward certain answers.   
Taking inspiration from \cite{zhao2021calibrate}, we have observed that the predictive bias also persists even when solely providing the template as input to the LLM. In Figure \ref{fig:template_bias}, the red line illustrates the bias associated with the template we employed in SST-2. Notably, when presented with context-free input, the LLM exhibits a tendency for a positive prediction with the possibility over 90\%. We refer to this bias as Template Bias \footnote{We use the term "Template Bias" to distinguish it from the Bias discussed in \cite{zhao2021calibrate}.} (See Appendix \ref{app_bias} for more details). Essentially, it characterizes the inherent bias present in the zero-shot prompt, whereby the mere utilization of a template prompts the LLM to generate predictions that favor specific answers, in the absence of any demonstration.

% \textcolor[rgb]{1,0,0}{$\bullet$}

\subsection{Calibration Before Sampling}
\label{cbs}
% We notice that the template bias may lead to an unfair evaluation of the IG when sampling examples. For instance, the data example around the red line in Figure \ref{fig:template_bias}, contains similar information to content-free input but has high IG (low conditional entropy). 
We have observed that the presence of template bias may result in an unfair evaluation of the IG when sampling examples. For instance, the data example located near the red line in Figure \ref{fig:template_bias}, exhibits similar information to content-free input but demonstrates high IG (low conditional entropy). 

% Therefore, to mitigate this template bias, we propose Calibration Before Sampling. We take the technique of applying the linear transformation\footnote{We applied the transformation on the output probabilities though it is often applied to logits. The reason is that we have only access to the output probabilities of GPT-3 via OpenAI API. To keep the same setting, we do the same for GPT-2 XL and GPT-J. }~\cite{platt1999probabilistic,guo2017calibration} to the output probabilities $\mathbf{p} = p_{\theta}(y|x_{ob}, T)$ to obtain the new calibrated probabilities $\mathbf{q} = q_{\theta}(y|x_{ob}, T)$ . That is,
To mitigate this template bias, we propose Calibration Before Sampling strategy\footnote{We refer to the contextual calibration method in \cite{zhao2021calibrate} as post-calibration. Calibration alone in our paper refers to the one before sampling proposed in our work. }. This involves applying the linear transformation\footnote{We apply the transformation to the output probabilities, although it is typically used for logits. The reason is that we only have access to the output probabilities of GPT-3 via OpenAI API. To maintain consistency across different LLMs, we apply the same transformation for GPT-2 XL and GPT-J. }~\cite{platt1999probabilistic,guo2017calibration} to the output probabilities $\mathbf{p} = p_{\theta}(y|x_{ob}, T)$ to obtain calibrated probabilities $\mathbf{q} = q_{\theta}(y|x_{ob}, T)$ . That is,
\begin{equation}
   \label{eq:cal}
   \mathbf{q} = \mathbf{\sigma}(\mathbf{Wp}+\mathbf{b})
\end{equation}
% where $\sigma$ is the normalization function and the weight matrix $\mathbf{W}$ is restricted to be a diagonal matrix known as vector scaling. We then estimate the $\mathbf{W}$ using the content-free strategy~\cite{zhao2021calibrate}. We construct the zero-shot prompt using the task-specific template $T$ and a set of content-free strings $\mathcal{D}_{cf}$ including the empty string, "N/A" and "[MASK]". We take the average of all output probabilities with each content-free string as input followed by normalization $\sigma$. 
where $\sigma$ denotes the normalization function, and the weight matrix $\mathbf{W}$ is constrained to be a diagonal matrix, known as vector scaling. To estimate $\mathbf{W}$, we leverage the content-free strategy~\cite{zhao2021calibrate}. We construct the zero-shot prompt using the task-specific template $T$ and a set of content-free strings $\mathcal{D}_{cf}$, which includes the empty string, "N/A" and "[MASK]". By averaging the output probabilities obtained from each content-free string as input, followed by normalization $\sigma$, we obtain, 
\begin{equation}
    \label{eq:cf}
    \mathbf{p_{cf}} = \sigma(\frac{1}{|\mathcal{D}_{cf}|}\sum_{x_{cf}\in \mathcal{D}_{cf}}p_{\theta}(y|x_{cf}, T))
\end{equation}
% As a consequence, to correct the bias from the template, we set $\mathbf{W} = \mathbf{diag}(\mathbf{p_{cf}})^{-1}$ and $\mathbf{b}$ to be zero vector. Then the calibrated conditional entropy can be eventually computed as
Consequently, to correct the bias stemming from the template, we set $\mathbf{W} = \mathbf{diag}(\mathbf{p_{cf}})^{-1}$ and $\mathbf{b}$ to be the zero vector. The calibrated conditional entropy can be then computed as,
\begin{equation}
    \label{eq:ce}
    H(Y|x_{ob}) = - \sum_{y\in Y} q_{\theta}(y|x_{ob}, T)\log q_{\theta}(y|x_{ob}, T)
\end{equation}
% The example in Figure \ref{fig:template_bias} demonstrates that after calibration the data sample around the red line shifts to a balanced place with low IG (high conditional entropy). Our algorithm is summarized as follows. 
The example depicted in Figure \ref{fig:template_bias} demonstrates that after calibration, the data sample located around the red line shifts to a position near the balanced line with low IG (high conditional entropy). 

The algorithm of our proposed method can be summarized as follows. 
\begin{algorithm}
\caption{Maximum Information Gain Sampling with Calibration Before Sampling}
\label{alg}
\begin{algorithmic}[1] 
\REQUIRE ~~\\ unlabeled dataset $\mathcal{D}_{unlab}$, number of examples to be sampled $K$, task-specific template $T$, LLM $\theta$
    \FOR{$x$ in $\mathcal{D}_{unlab}$}
        \STATE Construct prompt for $x$ using $T$
        \STATE Calculate $p_{\theta}(y|x, T)$ via LLM
        \STATE Calculate $\mathbf{p_{cf}}$ using Eq.\ref{eq:cf} 
        \STATE Calculate $\mathbf{q}$ via calibrating $\mathbf{p}$ using Eq.\ref{eq:cal}
        \STATE Evaluate IG via calculating H(Y|x) using Eq.\ref{eq:ce}
    \ENDFOR
 \STATE Rank all examples in $\mathcal{D}_{unlab}$ based on IG
\ENSURE ~~\\ Examples $\{x_j\}^K_{j=1}$ with top $K$ highest IG; 
\end{algorithmic}
\end{algorithm}

\section{Experimental Setup}

\begin{table*}[htbp]
\begin{center}
\begin{tabular}{cllcccccccccccccccccccccccc}
\toprule[1pt]
\multicolumn{3}{c}{LM} & \multicolumn{3}{c}{Method} & \multicolumn{3}{c}{SST-2} & \multicolumn{3}{c}{AGNews} & \multicolumn{3}{c}{TREC} & \multicolumn{3}{c}{CB} & \multicolumn{3}{c}{RTE} & \multicolumn{3}{c}{DBPedia} & \multicolumn{3}{c}{Avg} \\ \hline 

% gpt2 xl
\multicolumn{3}{c}{\multirow{4}{*}{GPT-2 1.5B}} & \multicolumn{3}{c}{Random} & \multicolumn{3}{c}{$59.7_{13.2}$} & \multicolumn{3}{c}{$39.6_{10.3}$} & \multicolumn{3}{c}{$27.1_{5.9}$} & \multicolumn{3}{c}{$28.6_{16.1}$} & \multicolumn{3}{c}{$53.4_{1.1}$} & \multicolumn{3}{c}{$51.2_{14.1}$} & \multicolumn{3}{c}{43.3} \\ 
\multicolumn{3}{c}{} & \multicolumn{3}{c}{MaxEntropy} & \multicolumn{3}{c}{$72.2_{10.4}$} & \multicolumn{3}{c}{$42.5_{8.1}$} & \multicolumn{3}{c}{$25.7_{6.5}$} & \multicolumn{3}{c}{$25.0_{19.8}$} & \multicolumn{3}{c}{$52.6_{1.4}$} & \multicolumn{3}{c}{$30.7_{14.1}$} & \multicolumn{3}{c}{41.5} \\ 
\multicolumn{3}{c}{} & \multicolumn{3}{c}{MaxIG} & \multicolumn{3}{c}{$52.8_{0.7}$} & \multicolumn{3}{c}{$\textbf{44.5}_{12.6}$} & \multicolumn{3}{c}{$\textbf{29.8}_{3.7}$} & \multicolumn{3}{c}{$\textbf{39.6}_{1.3}$} & \multicolumn{3}{c}{$\textbf{54.4}_{0.6}$} & \multicolumn{3}{c}{$\textbf{56.3}_{10.9}$} & \multicolumn{3}{c}{46.2} \\
\multicolumn{3}{c}{} & \multicolumn{3}{c}{CBS MaxIG} & \multicolumn{3}{c}{$\textbf{85.8}_{2.3}$} & \multicolumn{3}{c}{$32.9_{12.4}$} & \multicolumn{3}{c}{$29.7_{5.0}$} & \multicolumn{3}{c}{$\textbf{39.6}_{1.3}$} & \multicolumn{3}{c}{$53.9_{1.0}$} & \multicolumn{3}{c}{$50.8_{18.5}$} & \multicolumn{3}{c}{\textbf{48.8}} \\

% % ada
% \midrule[0.5pt]
% \multicolumn{3}{c}{\multirow{4}{*}{GPT-3 2.7B}} & \multicolumn{3}{c}{Random} & \multicolumn{3}{c}{$60.6_{8.9}$} & \multicolumn{3}{c}{$36.6_{6.4}$} & \multicolumn{3}{c}{$26.7_{6.6}$} & \multicolumn{3}{c}{$36.8_{22.9}$} & \multicolumn{3}{c}{$49.0_{2.6}$} & \multicolumn{3}{c}{$\textbf{22.7}_{9.4}$} & \multicolumn{3}{c}{38.7} \\ 
% \multicolumn{3}{c}{} & \multicolumn{3}{c}{MaxEntropy} & \multicolumn{3}{c}{$60.3_{5.5}$} & \multicolumn{3}{c}{$31.5_{10.7}$} & \multicolumn{3}{c}{$24.5_{3.8}$} & \multicolumn{3}{c}{$43.9_{17.6}$} & \multicolumn{3}{c}{$\textbf{53.9}_{1.7}$} & \multicolumn{3}{c}{$17.8_{6.2}$} & \multicolumn{3}{c}{38.7} \\ 
% \multicolumn{3}{c}{} & \multicolumn{3}{c}{MaxIG} & \multicolumn{3}{c}{$73.2_{8.3}$} & \multicolumn{3}{c}{$\textbf{41.1}_{8.3}$} & \multicolumn{3}{c}{$\textbf{32.5}_{7.1}$} & \multicolumn{3}{c}{$52.9_{3.5}$} & \multicolumn{3}{c}{$49.7_{1.9}$} & \multicolumn{3}{c}{$18.7_{5.7}$} & \multicolumn{3}{c}{\textbf{44.7}} \\
% \multicolumn{3}{c}{} & \multicolumn{3}{c}{Cal MaxIG} & \multicolumn{3}{c}{$\textbf{73.4}_{8.6}$} & \multicolumn{3}{c}{$27.3_{2.5}$} & \multicolumn{3}{c}{$27.9_{4.7}$} & \multicolumn{3}{c}{$\textbf{57.1}_{0.0}$} & \multicolumn{3}{c}{$52.9_{1.8}$} & \multicolumn{3}{c}{$12.6_{4.2}$} & \multicolumn{3}{c}{41.9} \\ 

% gptj
\midrule[0.5pt]
\multicolumn{3}{c}{\multirow{4}{*}{GPT-J 6B}} & \multicolumn{3}{c}{Random} & \multicolumn{3}{c}{$67.5_{6.6}$} & \multicolumn{3}{c}{$38.5_{13.8}$} & \multicolumn{3}{c}{$38.3_{10.7}$} & \multicolumn{3}{c}{$23.2_{7.4}$} & \multicolumn{3}{c}{$51.6_{4.1}$} & \multicolumn{3}{c}{$61.8_{15.0}$} & \multicolumn{3}{c}{46.8} \\ 
\multicolumn{3}{c}{} & \multicolumn{3}{c}{MaxEntropy} & \multicolumn{3}{c}{$75.3_{6.8}$} & \multicolumn{3}{c}{$26.9_{4.4}$} & \multicolumn{3}{c}{$38.4_{4.5}$} & \multicolumn{3}{c}{$26.4_{3.3}$} & \multicolumn{3}{c}{$51.0_{4.8}$} & \multicolumn{3}{c}{$70.9_{10.8}$} & \multicolumn{3}{c}{48.2} \\ 
\multicolumn{3}{c}{} & \multicolumn{3}{c}{MaxIG} & \multicolumn{3}{c}{$65.5_{5.3}$} & \multicolumn{3}{c}{$35.5_{11.6}$} & \multicolumn{3}{c}{$34.3_{6.1}$} & \multicolumn{3}{c}{$31.8_{2.9}$} & \multicolumn{3}{c}{$\textbf{55.3}_{3.6}$} & \multicolumn{3}{c}{$64.1_{7.2}$} & \multicolumn{3}{c}{47.8} \\
\multicolumn{3}{c}{} & \multicolumn{3}{c}{CBS MaxIG} & \multicolumn{3}{c}{$\textbf{76.8}_{9.1}$} & \multicolumn{3}{c}{$\textbf{39.1}_{12.0}$} & \multicolumn{3}{c}{$\textbf{47.8}_{8.2}$} & \multicolumn{3}{c}{$\textbf{32.5}_{4.4}$} & \multicolumn{3}{c}{$\textbf{55.3}_{3.6}$} & \multicolumn{3}{c}{$\textbf{83.7}_{5.2}$} & \multicolumn{3}{c}{\textbf{55.9}} \\ 

% davinci
\midrule[0.5pt]
% sst-2, agnews, trec, cb, rte, dbpedia, avg
\multicolumn{3}{c}{\multirow{4}{*}{GPT-3 175B}} & \multicolumn{3}{c}{Random} & \multicolumn{3}{c}{$87.3_{3.7}$} & \multicolumn{3}{c}{$62.8_{0.8}$} & \multicolumn{3}{c}{$54.8_{1.8}$} & \multicolumn{3}{c}{$39.3_{30.4}$} &  \multicolumn{3}{c}{$56.5_{0.5}$} & \multicolumn{3}{c}{$79.7_{5.7}$} & \multicolumn{3}{c}{63.4} \\ 
\multicolumn{3}{c}{} & \multicolumn{3}{c}{MaxEntropy} & \multicolumn{3}{c}{$\textbf{96.5}_{0.5}$} & \multicolumn{3}{c}{$63.0_{0.0}$} & \multicolumn{3}{c}{$58.7_{2.0}$} & \multicolumn{3}{c}{$38.4_{2.7}$} & \multicolumn{3}{c}{$\textbf{62.8}_{0.7}$} & \multicolumn{3}{c}{$80.0_{0.7}$} & \multicolumn{3}{c}{66.6} \\ 
\multicolumn{3}{c}{} & \multicolumn{3}{c}{MaxIG} & \multicolumn{3}{c}{$92.5_{3.5}$} & \multicolumn{3}{c}{$72.3_{2.3}$} & \multicolumn{3}{c}{$62.7_{4.0}$} & \multicolumn{3}{c}{$\textbf{41.1}_{0.0}$} & \multicolumn{3}{c}{$59.8_{3.8}$} & \multicolumn{3}{c}{$81.5_{0.8}$} & \multicolumn{3}{c}{68.3}  \\
\multicolumn{3}{c}{} & \multicolumn{3}{c}{CBS MaxIG} & \multicolumn{3}{c}{$96.2_{0.2}$} & \multicolumn{3}{c}{$\textbf{72.7}_{1.3}$} & \multicolumn{3}{c}{$\textbf{64.3}_{2.0}$} & \multicolumn{3}{c}{$\textbf{41.1}_{0.0}$} & \multicolumn{3}{c}{$60.3_{3.3}$} & \multicolumn{3}{c}{$\textbf{87.3}_{0.7}$} & \multicolumn{3}{c}{\textbf{70.3}}  \\
\bottomrule[1pt]
\end{tabular}
\caption{Main results for one-shot learning. The last column shows the average accuracies across all tasks. We report the mean and standard deviation across different random seeds. The template of each task is fixed. We bold the best result among all selection methods for each task and each LLM. }.
\label{table:main}
\vspace{-5mm}
\end{center}
\end{table*}

% \begin{table*}[htbp]
% \begin{center}
% \begin{tabular}{cllcccccccccc}
% \toprule[1pt]
% \multicolumn{3}{c}{\multirow{2}{*}{Dataset}} & \multicolumn{3}{c}{Random} & \multicolumn{3}{c}{Max-Entropy} & \multicolumn{2}{c}{Max-IG} & \multicolumn{2}{c}{Cal Max-IG} \\ \cline{4-13} 
% \multicolumn{3}{c}{} & R & P  & F1 & A & E & G & dist-1 & dist-2 & dist-1 & dist-2 \\ \hline
% \multicolumn{3}{c}{AGNews} & 0.232 & 0.232 & 0.232 & 0.915 & 0.511 & 0.798 & 0.935 & 0.969 & 0.093 & 0.097 \\ 
% \multicolumn{3}{c}{TREC} & 0.270 & 0.270 & 0.270 & 0.907 & 0.495 & 0.774 & 0.747 & 0.806 & 0.075 & 0.081 \\ 
% \multicolumn{3}{c}{CB} & 0.265 & 0.222 & 0.242 & 0.923 & 0.543 & 0.811 & 0.938 & 0.973 & 0.177 & 0.222 \\
% \multicolumn{3}{c}{RTE} & 0.256 & 0.224 & 0.239 & 0.923 & 0.540 & 0.812 & \bf{0.947} & \bf{0.976} & 0.165 & 0.206 \\
% \multicolumn{3}{c}{SST-2} & 0.259 & 0.244 & 0.251 & 0.914 & 0.530 & 0.818 & 0.821 & 0.911 & 0.106 & 0.126 \\
% \multicolumn{3}{c}{DBPedia} & 0.372 & \bf{0.286} & 0.323 & \bf{0.952} & 0.591 & 0.853 & 0.754 & 0.892 & 0.313 & 0.597 \\
% \midrule[1pt]
% \multicolumn{3}{c}{Avg} & \bf{0.422} & 0.272 & \bf{0.331} & 0.948 & \bf{0.609} & \bf{0.868} & 0.852 & 0.941 & \bf{0.412} & \bf{0.742} \\ 
% \bottomrule[1pt]
% \end{tabular}
% \caption{Performance comparison on the DailyDialog dataset(R:Recall, P:Precision, A:Average, E:Extrema, G:Greedy)}
% \label{table:dailydiag}
% \end{center}
% \end{table*}

\textbf{Evaluation Datasets. }  We experiment on six text classification datasets including binary sentiment analysis SST-2~\cite{sst2}, 6-way question classification TREC~\cite{trec}, 3-way textual entailment CB~\cite{cb}, RTE~\cite{rte} from SuperGLUE~\cite{superglue}, 4-way topic classification AGNews~\cite{agnews}, and 14-way DBPedia~\cite{agnews}. We use a fixed template (prompt format) for each dataset as per \cite{zhao2021calibrate}. Detailed information regarding each dataset can be found in Appendix \ref{app_dataset}. 
\vspace{3mm}

% \noindent \textbf{Models. }  We conduct our experiments on GPT-2 XL (1.5B parameters), GPT-J~\cite{wang2021gpt} (6B parameters), and GPT-3 davinci (175B parameters) for testing our method across various sizes of models. We get access to GPT-3 by OpenAI API.
\noindent \textbf{Models. }  For our experiments, we employ three distinct LLMs with different sizes: GPT-2 XL (1.5B parameters), GPT-J~\cite{wang2021gpt} (6B parameters), and GPT-3 davinci~\cite{brown2020language} (175B parameters). We get access to GPT-3 by using OpenAI API.
\vspace{3mm}

% \noindent \textbf{Baselines. }  Besides the \textbf{Random} baseline that randomly selects demonstration examples, we also include the commonly used uncertainty-based baseline \textbf{MaxEntropy} in active learning (AL)~\cite{dagan1995committee,settles2009active}, which greedily selects the demonstration example with maximum conditional entropy. The sampling objective of \textbf{MaxEntropy} is contrary to our proposed method. We denote our raw IG-based method by \textbf{MaxIG} and the variant using Calibration Before Sampling (CBS) as \textbf{CBS MaxIG}.
% Diversity-based baselines are excluded for our experiment setting deals with 1-shot, which is not suitable for diversity-based sampling. 

\noindent \textbf{Baselines. }  In addition to the \textbf{Random} baseline, which randomly selects demonstration examples, we also incorporate the widely utilized uncertainty-based baseline \textbf{MaxEntropy} in active learning (AL)~\cite{dagan1995committee,settles2009active}. The \textbf{MaxEntropy} baseline greedily selects the demonstration example with the highest conditional entropy. The sampling objective of \textbf{MaxEntropy} is opposite to that of our proposed method. We refer to our initial IG-based method as \textbf{MaxIG}, while the variant that incorporates Calibration Before Sampling (CBS) is denoted as \textbf{CBS MaxIG}.
\vspace{3mm}

% \noindent \textbf{Other Details. } In order to control the inference cost when sampling, we do not evaluate all data in the original training set. Instead, we first random sub-sample $N=100$ examples from the original training set as our $\mathcal{D}_{unlab}$ in our experiments. Then we evaluate all examples in $\mathcal{D}_{unlab}$ and sample from it. For each experiment using GPT-2 XL and GPT-J, we report the results using five different random seeds. For each experiment using GPT-3 davinci, we report results using two different random seeds. For evaluation, we sub-sample 300 samples of the test sets for all datasets as per~\cite{zhao2021calibrate,lu-etal-2022-fantastically} due to limited resources. 

\noindent \textbf{Other Details. } We use the original ICL, namely the \textit{direct} method, for all experiments in our work. In order to manage the inference cost during sampling, we do not evaluate the entire original training set. Instead, we first randomly sub-sample $N=100$ examples from the original training set to form the $\mathcal{D}_{unlab}$ in our experiments. Subsequently, we evaluate all the examples within $\mathcal{D}_{unlab}$ and perform sampling from this subset. For each experiment involving GPT-2 XL and GPT-J, we report the results based on five different random seeds. For experiments involving GPT-3 davinci, we report results using two different random seeds. For evaluation, we sub-sample 300 samples of the test sets for all datasets as per~\cite{zhao2021calibrate,lu-etal-2022-fantastically} due to limited resources. 

% Our main experiments are carried out in the one-shot setting. This is to avoid other potential factors such as ordering that may also influence the performance. In one-shot learning, we are able to focus on the impact of data informative ability. 

\section{Results}

\begin{figure*}[htbp]
    \centering
    \includegraphics[scale=0.39]{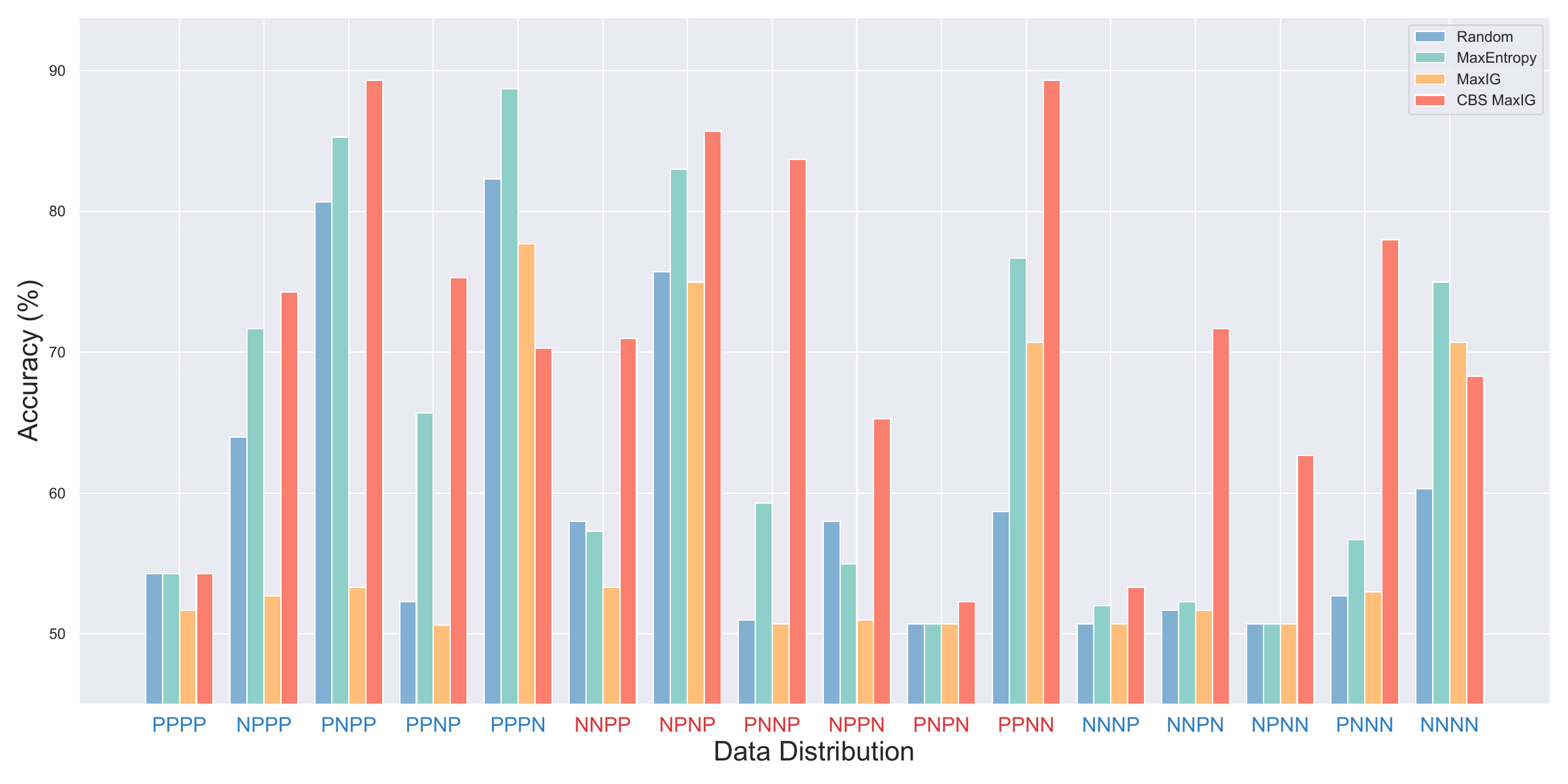}
    \caption{Four-shot performance comparison on SST-2 across different selection methods for different class balances and permutations. [P P N N] denotes two positive examples followed by two negative examples. We display the results for a total of 16 distinct types of four-shot demonstrations. Within this set, 6 types are balanced classes (highlighted in red), while the remaining 10 types are unbalanced classes (highlighted in blue).}
    \label{fig:4shot}
    \vspace{-2mm}
\end{figure*}
 
\subsection{Main Results}

% \noindent \textbf{One-shot Learning. } Our main experiments are carried out in the one-shot learning setting. This is to avoid other potential factors such as ordering that may also influence the performance. As a consequence, performing one-shot learning contributes to focusing on the impact of data informative ability on ICL performance. The main results are shown in Table \ref{table:main}. In terms of the average accuracies, our proposed CBS MaxIG performs the best and achieves relative improvements by 12.7\%, 19.4\%, 10.9\% for GPT-2 XL, GPT-J, and GPT-3 davinci compared to the random baseline, indicating the effectiveness of our proposed method. In comparison with MaxIG, the Calibration Before Sampling strategy improves tasks for GPT-J and GPT-3 davinci. This suggests that this strategy works better for larger models. Compared to Random and MaxEntropy baselines, our proposed MaxIG-based methods generally outperform them suggesting the effectiveness of sampling examples with maximum information gain and rationality of utilizing information gain for evaluating the data informative ability. 
\noindent \textbf{One-shot Learning. } Our main experiments are conducted in the case of one-shot learning to mitigate potential confounding factors such as the ordering of demonstrations that could influence performance outcomes. By focusing on one-shot learning, we aim to isolate the impact of data informative ability on the performance of ICL. The main results are presented in Table \ref{table:main}. In terms of the average accuracies, our proposed method, CBS MaxIG, exhibits superior performances, achieving relative improvements of 12.7\%, 19.4\%, 10.9\% over the random baseline for GPT-2 XL, GPT-J, and GPT-3 davinci, respectively. These results underscore the effectiveness of our proposed CBS MaxIG. Furthermore, when compared to the MaxIG approach, the Calibration Before Sampling strategy yields improvements in performances for GPT-J and GPT-3 davinci, suggesting that this strategy is particularly beneficial for larger models. Notably, our MaxIG-based methods consistently outperform the Random and MaxEntropy baselines, demonstrating the effectiveness of sampling examples with maximum IG and substantiating the validity of employing IG as a metric for evaluating data informative ability. 
\vspace{3mm}
% \noindent \textbf{Four-shot Learning. } To further evaluate our few-shot learning method, we extended our analysis to the four-shot learning scenario using the SST-2 dataset, which has been commonly employed in prior studies. We considered all possible combinations of class balances and permutations for the four-shot case, which encompass varying label distributions and orders. For each type, data examples were selected using the Random, MaxEntropy, MaxIG, and CBS MaxIG methods. This experimental design allowed us to examine the impact of data informativeness, given that class balance and order were fixed. The results are presented in Figure \ref{fig:4shot}. We observed that our proposed CBS MaxIG consistently outperformed all baselines in 13 out of 16 cases. For the remaining cases (PPPP, PPPN, and NNNN), we speculate that factors other than data informativeness, such as the label distribution, may exert a stronger influence on performance. Exploring the comparative impact of these different factors is left as an avenue for future research. In summary, our experimental findings underscore the importance of data informativeness and demonstrate the efficacy of our method in selecting the most informative data to enhance performance in scenarios involving more than one demonstration.

\noindent \textbf{Four-shot Learning. } To further evaluate our method for few-shot learning, we extend our experiments to the four-shot learning scenario on SST-2 using GPT-2 XL. We consider all possible combinations of class balances and permutations for the four-shot case\footnote{To sample data from different classes, we assume we have access to the target labels of the training set in this experiment. In our initial problem setting, we do not need the target labels during the sampling process.}, which encompass varying label distributions and orders. For each type, data examples are selected using Random, MaxEntropy, MaxIG, and CBS MaxIG methods respectively. This experimental design allows us to examine the impact of data informative ability, given that the class balance and order are fixed for one specific type. The results are presented in Figure \ref{fig:4shot}. We observe that our proposed CBS MaxIG consistently outperforms all baselines in 13 out of 16 cases. For the remaining cases (PPPP, PPPN, and NNNN), we conjecture that factors other than data informative ability, such as the label distribution, may exert a stronger influence on the performance. We leave comparing the impact of these different factors as future work. Overall, our experimental findings underscore the significance of the data informative ability and demonstrate the efficacy of our method in selecting the most informative data to enhance performance in scenarios involving more than one demonstration.

% \subsection{Combination with Existing Methods}
% To demonstrate that our method is orthogonal to the post calibration method~\cite{zhao2021calibrate} and the order probing method~\cite{lu-etal-2022-fantastically}, and can work together with them for better performance, we conduct experiments on SST-2 (binary classification) and DBPedia (14-classification) for comparative study.
% \vspace{3mm}

\subsection{Integration with Existing Methods}
In order to demonstrate that our method is orthogonal to prevailing techniques such as the post-calibration method and the order probing method, and illustrate their collaborative potential with our approach, we conduct experiments on two datasets, namely SST-2 (binary classification) and DBPedia (14-classification), for a comparative analysis.
\vspace{3mm}

% \noindent \textbf{Combination with Post-Calibration. } We compare the performances of Random and CBS MaxIG using post-calibration on one-shot learning across three LLMs. The results are shown in Table \ref{table:post_cal}. We find that using post-calibration on the selected examples using CBS MaxIG can achieve better performance than those randomly selected across different sizes of models. In addition, we find that our method without post-calibration can achieve similar or even outperform the post-calibration counterparts for GPT-3 davinci, indicating the effectiveness of our method. 
% \vspace{3mm}

\noindent \textbf{Integration with Post-Calibration. } To assess the performance of our method in conjunction with post-calibration, we compare the outcomes of Random and CBS MaxIG approaches on one-shot learning across three LLMs. The results, presented in Table \ref{table:post_cal}, reveal that by employing post-calibration on the selected examples using CBS MaxIG, superior performance is achieved compared to random selection, across different model sizes. Furthermore, it is observed that our method without post-calibration achieves comparable or even superior results to the post-calibration counterpart specifically for GPT-3 davinci, thereby affirming the effectiveness of our proposed approach.  
% \vspace{3mm}
\begin{table}[tbp]
\begin{center}
\begin{tabular}{lllllcccccccccc}
\toprule[1pt]
\multicolumn{5}{c}{} & \multicolumn{5}{c}{SST-2} & \multicolumn{5}{c}{DBPedia}  \\ \hline
% gpt2 xl
\multicolumn{5}{l}{\textit{GPT-2 1.5B}} & \multicolumn{5}{c}{} & \multicolumn{5}{c}{}  \\ 
\multicolumn{5}{l}{\;\;Random (\textbf{C})}  & \multicolumn{5}{c}{$76.7_{1.8}$} & \multicolumn{5}{c}{$68.3_{8.7}$} \\ 
\multicolumn{5}{l}{\;\;CBS MaxIG}  & \multicolumn{5}{c}{$\textbf{85.8}_{2.3}$} & \multicolumn{5}{c}{$50.8_{18.5}$} \\ 
\multicolumn{5}{l}{\;\;CBS MaxIG (\textbf{C})}  & \multicolumn{5}{c}{$74.5_{5.9}$} & \multicolumn{5}{c}{$\textbf{73.5}_{10.5}$} \\ 
\midrule[0.5pt]
% gptj 
\multicolumn{5}{l}{\textit{GPT-J 6B}} & \multicolumn{5}{c}{} & \multicolumn{5}{c}{}  \\ 
\multicolumn{5}{l}{\;\;Random (\textbf{C})}  & \multicolumn{5}{c}{$88.6_{1.6}$} & \multicolumn{5}{c}{$76.9_{3.4}$} \\ 
\multicolumn{5}{l}{\;\;CBS MaxIG}  & \multicolumn{5}{c}{$76.8_{9.1}$} & \multicolumn{5}{c}{$83.7_{5.2}$} \\ 
\multicolumn{5}{l}{\;\;CBS MaxIG (\textbf{C})}  & \multicolumn{5}{c}{$\textbf{94.5}_{0.6}$} & \multicolumn{5}{c}{$\textbf{86.7}_{0.8}$} \\ 
\midrule[0.5pt]
% gpt3 davinci 
\multicolumn{5}{l}{\textit{GPT-3 175B}} & \multicolumn{5}{c}{} & \multicolumn{5}{c}{}  \\ 
\multicolumn{5}{l}{\;\;Random (\textbf{C})}  & \multicolumn{5}{c}{$95.8_{0.8}$} & \multicolumn{5}{c}{$83.5_{3.5}$} \\ 
\multicolumn{5}{l}{\;\;CBS MaxIG}  & \multicolumn{5}{c}{$\textbf{96.2}_{0.2}$} & \multicolumn{5}{c}{$\textbf{87.3}_{0.7}$} \\ 
\multicolumn{5}{l}{\;\;CBS MaxIG (\textbf{C})}  & \multicolumn{5}{c}{$\textbf{96.2}_{0.2}$} & \multicolumn{5}{c}{$86.5_{0.8}$} \\

\bottomrule[1pt]
\end{tabular}
\caption{One-shot performance using post calibration. \textbf{C} denotes the sampling method using post-calibration.}
\label{table:post_cal}
\end{center}
\vspace{-4mm}
\end{table}

\begin{figure}[htbp]
  \centering
  \includegraphics[scale=0.3]{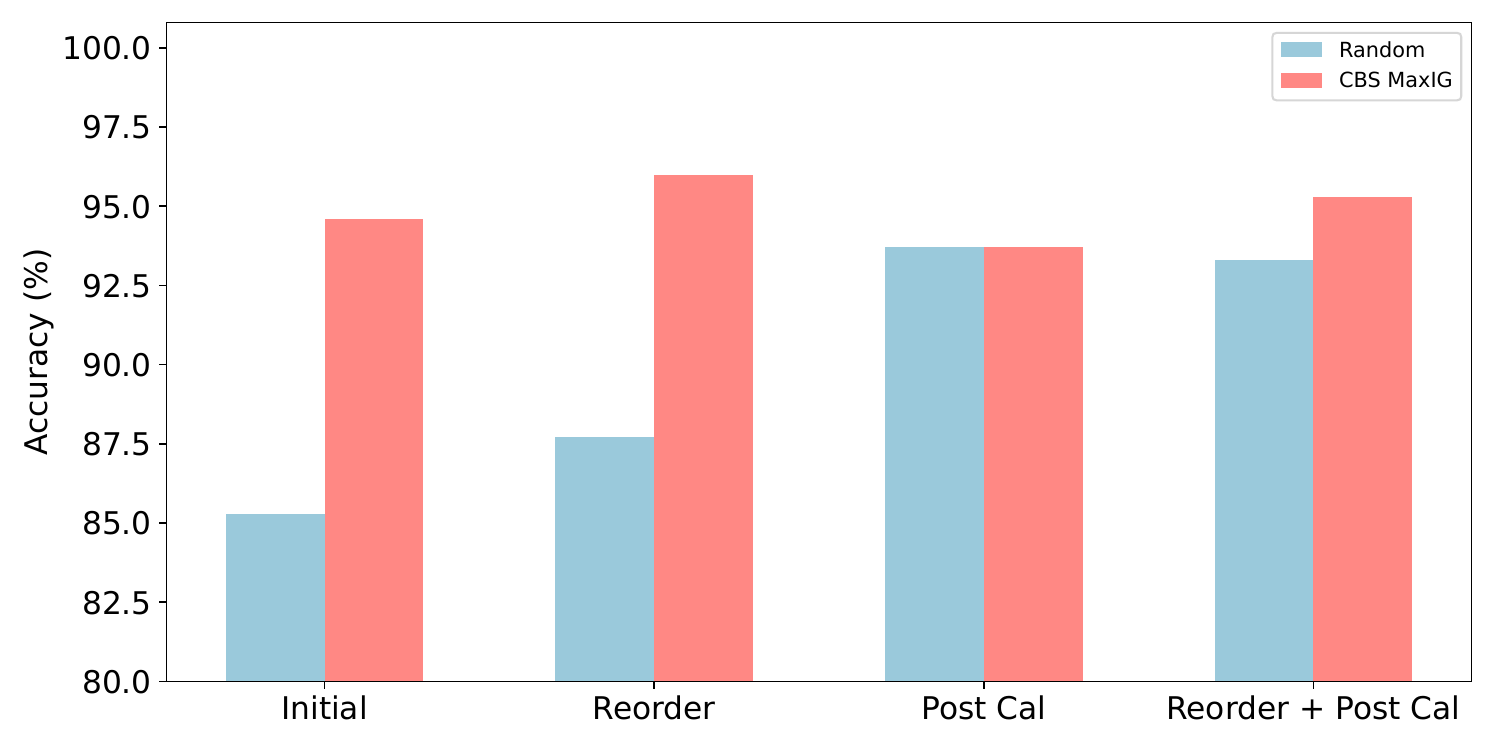}
  \caption{Four-shot performance comparison of four-shot learning on SST-2 using GPT-J. "Initial" represents the original one. "Reorder" denotes the order probing method, and "Post Cal" indicates the post-calibration method. "Reorder + Post Cal" represents the combination of order probing followed by post-calibration.}
  \label{fig:ord+cal}
  \vspace{-3mm}
\end{figure}

% \noindent \textbf{Combination with Order Probing. } We compare the performances of Random and CBS MaxIG when combined with order probing and post-calibration together for four-shot learning on GPT-J. We first sample four examples using Random and CBS MaxIG methods\footnote{We naively sample the examples with top-four highest IG without considering label distribution.}, respectively. Then we perform ordering probing and sample the permutation with maximum global entropy on the probing set. We present the results in Figure \ref{fig:ord+cal}. It is discovered that order probing can improve performance for both sampling methods. We also discover that order probing and post-calibration contribute more to the enhancement of the random baseline compared to our CBS MaxIG, suggesting our proposed method is more robust to the order and bias than random baseline. 

\noindent \textbf{Integration with Order Probing. } To assess the performance of Random and CBS MaxIG methods in conjunction with order probing and post-calibration for four-shot learning using GPT-J, we first sample four examples using Random and CBS MaxIG methods\footnote{Note that we naively sample the examples with the top-four highest IG without considering label distribution}, respectively. Subsequently, we perform ordering probing and sample the permutation with maximum global entropy on the probing set\footnote{We utilize the global entropy metric due to its superior performance in the original paper~\cite{lu-etal-2022-fantastically}.}. The results, depicted in Figure \ref{fig:ord+cal}, reveal that ordering probing improves the performance for both sampling methods. Furthermore, it is discovered that order probing and post-calibration contribute more significantly to enhancing the performance of the Random baseline compared to our CBS MaxIG approach, thereby suggesting that our proposed method is more robust to order and bias factors in comparison to the Random baseline. 

\begin{table}[tbp]
\begin{center}
\begin{tabular}{lllllcccccccccc}
\toprule[1pt]
\multicolumn{5}{c}{} & \multicolumn{5}{c}{SST-2} & \multicolumn{5}{c}{DBPedia}  \\ \hline
% gpt2 xl
\multicolumn{5}{l}{\textit{GPT-2 1.5B}} & \multicolumn{5}{c}{} & \multicolumn{5}{c}{}  \\ 
\multicolumn{5}{l}{\;\;MaxEntropy}  & \multicolumn{5}{c}{$\textbf{72.2}_{10.4}$} & \multicolumn{5}{c}{$30.7_{14.1}$} \\ 
\multicolumn{5}{l}{\;\;CBS MaxEntropy}  & \multicolumn{5}{c}{$53.1_{1.7}$} & \multicolumn{5}{c}{$\textbf{32.0}_{13.0}$} \\ 
\midrule[0.5pt]
% gptj 
\multicolumn{5}{l}{\textit{GPT-J 6B}} & \multicolumn{5}{c}{} & \multicolumn{5}{c}{}  \\  
\multicolumn{5}{l}{\;\;MaxEntropy}  & \multicolumn{5}{c}{$\textbf{75.3}_{6.8}$} & \multicolumn{5}{c}{$\textbf{70.9}_{10.8}$} \\ 
\multicolumn{5}{l}{\;\;CBS MaxEntropy}  & \multicolumn{5}{c}{$73.5_{4.6}$} & \multicolumn{5}{c}{$54.9_{14.2}$} \\ 
\midrule[0.5pt]
% gpt3 davinci 
\multicolumn{5}{l}{\textit{GPT-3 175B}} & \multicolumn{5}{c}{} & \multicolumn{5}{c}{}  \\ 
\multicolumn{5}{l}{\;\;MaxEntropy}  & \multicolumn{5}{c}{$\textbf{96.5}_{0.5}$} & \multicolumn{5}{c}{$\textbf{80.0}_{0.7}$} \\ 
\multicolumn{5}{l}{\;\;CBS MaxEntropy}  & \multicolumn{5}{c}{$88.2_{3.8}$} & \multicolumn{5}{c}{$78.2_{3.8}$} \\

\bottomrule[1pt]
\end{tabular}
\caption{Ablation on Calibration Before Sampling for MaxEntropy in one-shot learning.}
\label{table:abl_cbs}
\end{center}
\vspace{-5mm}
\end{table}

\subsection{Ablation on Calibration Before Sampling}

% Although the results in Table \ref{table:main} demonstrate the effectiveness of our Calibration Before Sampling strategy, we notice that the evaluation of MaxEntropy may also be biased by the template because the calculation of conditional entropy involves output distribution of LLM. As such, to further facilitate a fair assessment, we take the same calibration before sampling on MaxEntropy and report the results in Table \ref{table:abl_cbs}. It is discovered that there are significant performance drops across three LLMs with an exception for DBPedia using GPT-2 XL. The reason for the performance drop is that examples with maximum conditional entropy before calibration may have high IG after calibration. On the contrary, CBS MaxEntropy selects those with maximum entropy after calibration despite their high IG before calibration. We point it out because this further indicates that our proposed MaxIG-based sampling is more effective compared to the MaxEntropy-based method. This provides an insight that for ICL where parameters are not updated, MaxEntropy proposed in the traditional AL that needs to update parameters is not applicable. Certain examples with high IG contribute more to the ICL than uncertain examples with high entropy. This may be brought by the setting difference between ICL and traditional AL and how the difference influences the choice of sampling strategy needs to be investigated further in future work. 
Although the results in Table \ref{table:main} demonstrate the effectiveness of our Calibration Before Sampling strategy, it should be noted that the evaluation of the MaxEntropy method may also be subject to bias introduced by the template utilized, as the calculation of conditional entropy relies on the output distribution of the LLM. To ensure a fair assessment, we apply the same Calibration Before Sampling strategy to the MaxEntropy method, named CBS MaxEntropy, and report the corresponding one-shot outcomes of SST-2 and DBPedia in Table \ref{table:abl_cbs}. Notably, a significant decline in performance is observed across all three LLMs, with the exception of DBPedia when employing GPT-2 XL. The performance degradation can be attributed to the fact that examples selected by MaxEntropy may not have the highest entropy. Instead, these examples could correspond to the ones with high IG after calibration. Conversely, examples located near the Template Bias line in Figure \ref{fig:template_bias} are the ones with high entropy after calibration, and CBS MaxEntropy selects those examples. We emphasize this observation as it further reinforces the superiority of our proposed MaxIG-based sampling methods over the MaxEntropy-based approaches. This insight highlights that the MaxEntropy method from conventional active learning, which relies on parameter updates, is not perfectly suitable for ICL where parameters remain static. In such cases, certain examples with high IG contribute more significantly to ICL compared to uncertain examples with high entropy. The distinction between ICL and traditional active learning settings, and how this distinction influences the choice of sampling strategy, warrants further investigation in future work.   

% The performance degradation can be attributed to the fact that examples exhibiting maximum conditional entropy (MaxEntropy) prior to calibration may exhibit high IG after calibration. Conversely, CBS MaxEntropy selects examples with maximum entropy after calibration, irrespective of their initial IG. 

\begin{table}[tbp]
\begin{center}
\begin{tabular}{lllllccccccccccccccc}
\toprule[1pt]
\multicolumn{5}{l}{\;\;Method} & \multicolumn{5}{c}{Gold } & \multicolumn{5}{c}{Random } & \multicolumn{5}{c}{Drop(\%)}  \\ \hline

\multicolumn{5}{l}{\textit{SST-2}} & \multicolumn{5}{c}{} & \multicolumn{5}{c}{} & \multicolumn{5}{c}{}  \\ 
\multicolumn{5}{l}{\;\;Random }  & \multicolumn{5}{c}{$67.6_{6.6}$} & \multicolumn{5}{c}{$66.7_{8.5}$} & \multicolumn{5}{c}{$1.2\%$} \\ 
\multicolumn{5}{l}{\;\;CBS MaxIG} & \multicolumn{5}{c}{$76.8_{9.1}$} & \multicolumn{5}{c}{$67.4_{5.3}$} & \multicolumn{5}{c}{$\textbf{12.2\%}$} \\ 

\midrule[0.5pt]

\multicolumn{5}{l}{\textit{DBPedia}} & \multicolumn{5}{c}{} & \multicolumn{5}{c}{} & \multicolumn{5}{c}{}  \\ 
\multicolumn{5}{l}{\;\;Random }  & \multicolumn{5}{c}{$61.8_{15.0}$} & \multicolumn{5}{c}{$50.3_{8.1}$} & \multicolumn{5}{c}{$18.6\%$} \\ 
\multicolumn{5}{l}{\;\;CBS MaxIG} & \multicolumn{5}{c}{$83.7_{5.2}$} & \multicolumn{5}{c}{$36.3_{11.6}$} & \multicolumn{5}{c}{$\textbf{56.6\%}$} \\

\bottomrule[1pt]
\end{tabular}
\caption{One-shot performance using demonstrations with gold and random labels. The last column shows the percentage of performance drop.}
\label{table:gold}
\end{center}
\vspace{-5mm}
\end{table}

\section{Analysis }

\subsection{Gold Labels vs. Random Labels}
% Our sampling process from the unlabeled training set involves no target labels, we aim to study whether the highly informative data benefit from the golden label inspired from \cite{min-etal-2022-rethinking} in this experiment. We consider the demonstrations selected by Random baseline and our CBS MaxIG in one-shot learning using GPT-J. For each demonstration, we replace the gold labels with random labels and evaluate the performance respectively. The results are reported in Table \ref{table:gold}. We observe large performance drops when the gold labels are replaced by random labels for demonstrations selected by CBS MaxIG, while small drops for compared to those randomly selected. This suggests that informative data rely more on the correct labels. We conjecture that providing a wrong label for highly informative data may confuse the LLM and lower gain the information gain. 
Since our sampling process from the unlabeled training set involves no utilization of target labels, it is pertinent to examine whether the highly informative data derived from it can benefit from the gold labels, as inspired by \cite{min-etal-2022-rethinking}. In this experiment, we compare the performance of using demonstrations selected by the Random baseline and our CBS MaxIG approach in one-shot learning using GPT-J. For each demonstration, we replace the gold label with a random label from a small discrete set of possible labels, and evaluate the performance accordingly. The results are presented in Table \ref{table:gold}.

We observe a substantial decrease in performance when the gold labels are substituted with random labels for demonstrations selected by CBS MaxIG, whereas the drop is comparatively smaller for those randomly selected. This observation suggests that highly informative data heavily rely on the presence of accurate labels. We posit that providing incorrect labels for highly informative data may lead to confusion within the LLM and subsequently result in diminished information gain.  

% \subsection{Consistency over examples with high IG}
% To examine the consistency of performance for other examples with high IG, we select the examples with the top three largest IG and take them as the demonstration in one-shot learning separately. We plot the results for each of them in Figure. The experimental results demonstrate that demonstrations with high IG consistently outperform the random baselines, reinforcing the significance of MaxIG-based sampling. It is also observed that the demonstration with the third largest IG outperforms the one with the second largest IG. We believe this may come from the error of estimating the IG due to the used content-free strategy. 

% \begin{figure}[htbp]
%   \centering
%   \includegraphics[scale=0.5]{emnlp2023-latex/img/topk.pdf}
%   \caption{Performance of examples with Top-K highest IG on SST-2 and DBPedia using GPT-J.}
%   \label{fig:topk}
% \end{figure}

\subsection{Consistency of Examples with High IG}
\begin{figure}[tbp]
  \centering
  \includegraphics[scale=0.5]{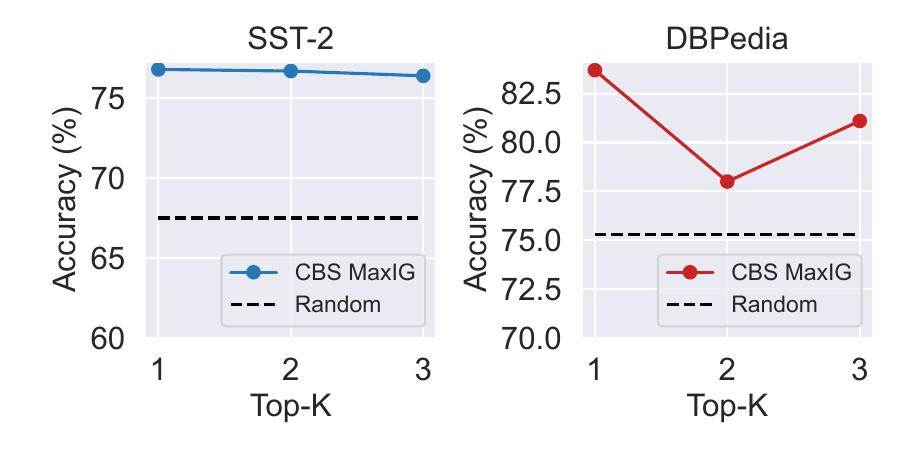}
  \caption{One-shot performance of examples with the Top-K highest IG on SST-2 and DBPedia using GPT-J.}
  \label{fig:topk}
  \vspace{-3mm}
\end{figure}

In order to assess the consistency of performance across other examples with high IG, we individually select the examples with the top-K highest IG values and utilize them as demonstrations in one-shot learning. The results for each selected example are presented in Figure \ref{fig:topk}, together with Random baseline depicted with dashed lines. The experimental findings demonstrate that demonstrations with high IG consistently outperform the Random baseline, thereby reinforcing the significance of employing MaxIG-based sampling. Notably, it is observed that the demonstration with the third-highest IG outperforms the one with the second-highest IG. We attribute this discrepancy to potential errors in estimating the IG, which may arise from the utilization of the content-free strategy. 

\section{Conclusion }
In this study, we have highlighted the significance of the data informative ability in ICL. We have demonstrated that data samples with varying informative abilities, even when subjected to the same input distribution, order and prompt formats, make distinct contributions to the overall performance of ICL. To address this, we draw inspiration from information theory and proposed to quantify the informative ability by evaluating the information gain of data samples. Moreover, we identify the presence of template bias, which could introduce unfairness in the evaluation of IG. To mitigate this bias, we introduce the Calibration Before Sampling strategy. Through extensive experiments, we have validated the effectiveness of the proposed maximum information gain sampling strategy and calibration before sampling strategy. This validation underscores the reliability of measuring informative ability based on information gain. Furthermore, our experimental findings illustrate that our method is orthogonal to existing approaches and can synergistically collaborate with them to achieve further performance improvements. We hope that our work can provide valuable guidance for the development of data-efficient methods and facilitate the exploration of enhanced data-centric approaches for ICL in the future.

\section*{Limitations}

% In this work, we only examine the text classification tasks in which the information gain is well defined on the prediction distribution with tractable conditional entropy. Future work can extend the experiments to generation tasks. Nevertheless, it is not trivial because a tractable definition of information gain on the output distribution with open words and various lengths is required. In addition, we do not consider the diversity of examples during the sampling process. Instead, we focus more on the data informative ability and conduct all experiments in one-shot and four-shot scenarios where diversity is not applicable in this case.    

% Our evaluation of information gain is model-aware and relies on the specific LLM, which indicates that the evaluation results cannot be shared across different models. So, when using a new model, we have to re-compute the information gain, which is not efficient enough and lead to additional computational cost. Besides, the computational cost also depends on the size of the training data pool because we evaluate each candidate in the pool. Although the parameters of LLMs do not need to be updated, multiple times of inferences still consume a large amount of computational resources especially when the LLM size is extremely large.

There are several limitations to consider in our work. Firstly, we focus solely on text classification tasks where the information gain can be well-defined based on the prediction distribution and tractable conditional entropy. Future research could extend our experiments to generation tasks. However, this extension poses challenges as it requires a tractable definition of information gain for output distributions that contain open words and have variable lengths.

Additionally, our sampling process does not explicitly consider the diversity of examples. Instead, we prioritize the data informative ability and conduct experiments in one-shot and four-shot scenarios where diversity is not as significant as in other cases with the goal of sampling many samples. Exploring methods to incorporate diversity during the sampling process is of importance for future work.   

Another limitation lies in the model-aware evaluation of information gain, which relies on the specific LLM used. This implies that the evaluation results cannot be directly applied to different models. When using a new model, the information gain for each example needs to be recomputed, which incurs additional computational cost. Moreover, the computational cost depends on the size of the training data pool, as each candidate example in the pool needs to be evaluated. Although the parameters of LLMs do not need to be updated, the repeated inferences still consume substantial computational resources, particularly when dealing with extremely large LMs.   

\section*{Acknowledgements }
We thank anonymous reviewers for their valuable feedback on the paper. We also thank Hengguan Huang and Yisong Miao for their helpful discussions.

% \section*{Ethics Statement}

% \section*{Acknowledgements}

\newpage
\bibliography{anthology,custom} 

\begin{thebibliography}{35}
\expandafter\ifx\csname natexlab\endcsname\relax\def\natexlab#1{#1}\fi

\bibitem[{Ash(2012)}]{ash2012information}
Robert~B Ash. 2012.
\newblock \emph{Information theory}.
\newblock Courier Corporation.

\bibitem[{Bowman et~al.(2015)Bowman, Angeli, Potts, and Manning}]{snli}
Samuel~R. Bowman, Gabor Angeli, Christopher Potts, and Christopher~D. Manning.
  2015.
\newblock \href {https://doi.org/10.18653/v1/D15-1075} {A large annotated
  corpus for learning natural language inference}.
\newblock In \emph{Proceedings of the 2015 Conference on Empirical Methods in
  Natural Language Processing}, pages 632--642, Lisbon, Portugal. Association
  for Computational Linguistics.

\bibitem[{Brown et~al.(2020)Brown, Mann, Ryder, Subbiah, Kaplan, Dhariwal,
  Neelakantan, Shyam, Sastry, Askell et~al.}]{brown2020language}
Tom Brown, Benjamin Mann, Nick Ryder, Melanie Subbiah, Jared~D Kaplan, Prafulla
  Dhariwal, Arvind Neelakantan, Pranav Shyam, Girish Sastry, Amanda Askell,
  et~al. 2020.
\newblock Language models are few-shot learners.
\newblock \emph{Advances in neural information processing systems},
  33:1877--1901.

\bibitem[{Chang and Jia(2022)}]{chang2022careful}
Ting-Yun Chang and Robin Jia. 2022.
\newblock Careful data curation stabilizes in-context learning.
\newblock \emph{arXiv preprint arXiv:2212.10378}.

\bibitem[{Clark et~al.(2019)Clark, Lee, Chang, Kwiatkowski, Collins, and
  Toutanova}]{boolq}
Christopher Clark, Kenton Lee, Ming-Wei Chang, Tom Kwiatkowski, Michael
  Collins, and Kristina Toutanova. 2019.
\newblock \href {https://doi.org/10.18653/v1/N19-1300} {{B}ool{Q}: Exploring
  the surprising difficulty of natural yes/no questions}.
\newblock In \emph{Proceedings of the 2019 Conference of the North {A}merican
  Chapter of the Association for Computational Linguistics: Human Language
  Technologies, Volume 1 (Long and Short Papers)}, pages 2924--2936,
  Minneapolis, Minnesota. Association for Computational Linguistics.

\bibitem[{Dagan and Engelson(1995)}]{dagan1995committee}
Ido Dagan and Sean~P Engelson. 1995.
\newblock Committee-based sampling for training probabilistic classifiers.
\newblock In \emph{Machine Learning Proceedings 1995}, pages 150--157.
  Elsevier.

\bibitem[{Dagan et~al.(2006)Dagan, Glickman, and Magnini}]{rte}
Ido Dagan, Oren Glickman, and Bernardo Magnini. 2006.
\newblock The pascal recognising textual entailment challenge.
\newblock In \emph{Machine Learning Challenges. Evaluating Predictive
  Uncertainty, Visual Object Classification, and Recognising Tectual
  Entailment: First PASCAL Machine Learning Challenges Workshop, MLCW 2005,
  Southampton, UK, April 11-13, 2005, Revised Selected Papers}, pages 177--190.
  Springer.

\bibitem[{De~Marneffe et~al.(2019)De~Marneffe, Simons, and Tonhauser}]{cb}
Marie-Catherine De~Marneffe, Mandy Simons, and Judith Tonhauser. 2019.
\newblock The commitmentbank: Investigating projection in naturally occurring
  discourse.
\newblock In \emph{proceedings of Sinn und Bedeutung}, volume~23, pages
  107--124.

\bibitem[{Devlin et~al.(2019)Devlin, Chang, Lee, and
  Toutanova}]{devlin-etal-2019-bert}
Jacob Devlin, Ming-Wei Chang, Kenton Lee, and Kristina Toutanova. 2019.
\newblock \href {https://doi.org/10.18653/v1/N19-1423} {{BERT}: Pre-training of
  deep bidirectional transformers for language understanding}.
\newblock In \emph{Proceedings of the 2019 Conference of the North {A}merican
  Chapter of the Association for Computational Linguistics: Human Language
  Technologies, Volume 1 (Long and Short Papers)}, pages 4171--4186,
  Minneapolis, Minnesota. Association for Computational Linguistics.

\bibitem[{Ein-Dor et~al.(2020)Ein-Dor, Halfon, Gera, Shnarch, Dankin, Choshen,
  Danilevsky, Aharonov, Katz, and Slonim}]{alemp}
Liat Ein-Dor, Alon Halfon, Ariel Gera, Eyal Shnarch, Lena Dankin, Leshem
  Choshen, Marina Danilevsky, Ranit Aharonov, Yoav Katz, and Noam Slonim. 2020.
\newblock \href {https://doi.org/10.18653/v1/2020.emnlp-main.638} {{A}ctive
  {L}earning for {BERT}: {A}n {E}mpirical {S}tudy}.
\newblock In \emph{Proceedings of the 2020 Conference on Empirical Methods in
  Natural Language Processing (EMNLP)}, pages 7949--7962, Online. Association
  for Computational Linguistics.

\bibitem[{Gao et~al.(2021)Gao, Fisch, and Chen}]{gao-etal-2021-making}
Tianyu Gao, Adam Fisch, and Danqi Chen. 2021.
\newblock \href {https://doi.org/10.18653/v1/2021.acl-long.295} {Making
  pre-trained language models better few-shot learners}.
\newblock In \emph{Proceedings of the 59th Annual Meeting of the Association
  for Computational Linguistics and the 11th International Joint Conference on
  Natural Language Processing (Volume 1: Long Papers)}, pages 3816--3830,
  Online. Association for Computational Linguistics.

\bibitem[{Guo et~al.(2017)Guo, Pleiss, Sun, and
  Weinberger}]{guo2017calibration}
Chuan Guo, Geoff Pleiss, Yu~Sun, and Kilian~Q Weinberger. 2017.
\newblock On calibration of modern neural networks.
\newblock In \emph{International conference on machine learning}, pages
  1321--1330. PMLR.

\bibitem[{Hongjin et~al.(2023)Hongjin, Kasai, Wu, Shi, Wang, Xin, Zhang,
  Ostendorf, Zettlemoyer, Smith et~al.}]{hongjin2023selective}
SU~Hongjin, Jungo Kasai, Chen~Henry Wu, Weijia Shi, Tianlu Wang, Jiayi Xin, Rui
  Zhang, Mari Ostendorf, Luke Zettlemoyer, Noah~A Smith, et~al. 2023.
\newblock Selective annotation makes language models better few-shot learners.
\newblock In \emph{The Eleventh International Conference on Learning
  Representations}.

\bibitem[{Li and Qiu(2023)}]{li2023finding}
Xiaonan Li and Xipeng Qiu. 2023.
\newblock Finding supporting examples for in-context learning.
\newblock \emph{arXiv preprint arXiv:2302.13539}.

\bibitem[{Liu et~al.(2021)Liu, Shen, Zhang, Dolan, Carin, and
  Chen}]{liu2021makes}
Jiachang Liu, Dinghan Shen, Yizhe Zhang, Bill Dolan, Lawrence Carin, and Weizhu
  Chen. 2021.
\newblock What makes good in-context examples for gpt-$3 $?
\newblock \emph{arXiv preprint arXiv:2101.06804}.

\bibitem[{Lu et~al.(2022)Lu, Bartolo, Moore, Riedel, and
  Stenetorp}]{lu-etal-2022-fantastically}
Yao Lu, Max Bartolo, Alastair Moore, Sebastian Riedel, and Pontus Stenetorp.
  2022.
\newblock \href {https://doi.org/10.18653/v1/2022.acl-long.556} {Fantastically
  ordered prompts and where to find them: Overcoming few-shot prompt order
  sensitivity}.
\newblock In \emph{Proceedings of the 60th Annual Meeting of the Association
  for Computational Linguistics (Volume 1: Long Papers)}, pages 8086--8098,
  Dublin, Ireland. Association for Computational Linguistics.

\bibitem[{Margatina et~al.(2021)Margatina, Vernikos, Barrault, and
  Aletras}]{contrastive}
Katerina Margatina, Giorgos Vernikos, Lo{\"\i}c Barrault, and Nikolaos Aletras.
  2021.
\newblock \href {https://doi.org/10.18653/v1/2021.emnlp-main.51} {Active
  learning by acquiring contrastive examples}.
\newblock In \emph{Proceedings of the 2021 Conference on Empirical Methods in
  Natural Language Processing}, pages 650--663, Online and Punta Cana,
  Dominican Republic. Association for Computational Linguistics.

\bibitem[{Min et~al.(2022)Min, Lyu, Holtzman, Artetxe, Lewis, Hajishirzi, and
  Zettlemoyer}]{min-etal-2022-rethinking}
Sewon Min, Xinxi Lyu, Ari Holtzman, Mikel Artetxe, Mike Lewis, Hannaneh
  Hajishirzi, and Luke Zettlemoyer. 2022.
\newblock \href {https://aclanthology.org/2022.emnlp-main.759} {Rethinking the
  role of demonstrations: What makes in-context learning work?}
\newblock In \emph{Proceedings of the 2022 Conference on Empirical Methods in
  Natural Language Processing}, pages 11048--11064, Abu Dhabi, United Arab
  Emirates. Association for Computational Linguistics.

\bibitem[{Perez et~al.(2021)Perez, Kiela, and Cho}]{perez2021true}
Ethan Perez, Douwe Kiela, and Kyunghyun Cho. 2021.
\newblock True few-shot learning with language models.
\newblock \emph{Advances in neural information processing systems},
  34:11054--11070.

\bibitem[{Platt et~al.(1999)}]{platt1999probabilistic}
John Platt et~al. 1999.
\newblock Probabilistic outputs for support vector machines and comparisons to
  regularized likelihood methods.
\newblock \emph{Advances in large margin classifiers}, 10(3):61--74.

\bibitem[{Rubin et~al.(2021)Rubin, Herzig, and Berant}]{rubin2021learning}
Ohad Rubin, Jonathan Herzig, and Jonathan Berant. 2021.
\newblock Learning to retrieve prompts for in-context learning.
\newblock \emph{arXiv preprint arXiv:2112.08633}.

\bibitem[{Settles(2009)}]{settles2009active}
Burr Settles. 2009.
\newblock Active learning literature survey.

\bibitem[{Socher et~al.(2013)Socher, Perelygin, Wu, Chuang, Manning, Ng, and
  Potts}]{sst2}
Richard Socher, Alex Perelygin, Jean Wu, Jason Chuang, Christopher~D. Manning,
  Andrew Ng, and Christopher Potts. 2013.
\newblock \href {https://aclanthology.org/D13-1170} {Recursive deep models for
  semantic compositionality over a sentiment treebank}.
\newblock In \emph{Proceedings of the 2013 Conference on Empirical Methods in
  Natural Language Processing}, pages 1631--1642, Seattle, Washington, USA.
  Association for Computational Linguistics.

\bibitem[{Voorhees and Tice(2000)}]{trec}
Ellen~M Voorhees and Dawn~M Tice. 2000.
\newblock Building a question answering test collection.
\newblock In \emph{Proceedings of the 23rd annual international ACM SIGIR
  conference on Research and development in information retrieval}, pages
  200--207.

\bibitem[{Wan et~al.(2023{\natexlab{a}})Wan, Sun, Dai, Arik, and
  Pfister}]{wanaclfinding}
Xingchen Wan, Ruoxi Sun, Hanjun Dai, Sercan Arik, and Tomas Pfister.
  2023{\natexlab{a}}.
\newblock \href {https://doi.org/10.18653/v1/2023.findings-acl.216} {Better
  zero-shot reasoning with self-adaptive prompting}.
\newblock In \emph{Findings of the Association for Computational Linguistics:
  ACL 2023}, pages 3493--3514, Toronto, Canada. Association for Computational
  Linguistics.

\bibitem[{Wan et~al.(2023{\natexlab{b}})Wan, Sun, Nakhost, Dai, Eisenschlos,
  Arik, and Pfister}]{wan2023universal}
Xingchen Wan, Ruoxi Sun, Hootan Nakhost, Hanjun Dai, Julian~Martin Eisenschlos,
  Sercan~O Arik, and Tomas Pfister. 2023{\natexlab{b}}.
\newblock Universal self-adaptive prompting.
\newblock \emph{arXiv preprint arXiv:2305.14926}.

\bibitem[{Wang et~al.(2019)Wang, Pruksachatkun, Nangia, Singh, Michael, Hill,
  Levy, and Bowman}]{superglue}
Alex Wang, Yada Pruksachatkun, Nikita Nangia, Amanpreet Singh, Julian Michael,
  Felix Hill, Omer Levy, and Samuel Bowman. 2019.
\newblock Superglue: A stickier benchmark for general-purpose language
  understanding systems.
\newblock \emph{Advances in neural information processing systems}, 32.

\bibitem[{Wang and Komatsuzaki(2021)}]{wang2021gpt}
Ben Wang and Aran Komatsuzaki. 2021.
\newblock Gpt-j-6b: A 6 billion parameter autoregressive language model.

\bibitem[{Williams et~al.(2018)Williams, Nangia, and Bowman}]{mnli}
Adina Williams, Nikita Nangia, and Samuel Bowman. 2018.
\newblock \href {https://doi.org/10.18653/v1/N18-1101} {A broad-coverage
  challenge corpus for sentence understanding through inference}.
\newblock In \emph{Proceedings of the 2018 Conference of the North {A}merican
  Chapter of the Association for Computational Linguistics: Human Language
  Technologies, Volume 1 (Long Papers)}, pages 1112--1122, New Orleans,
  Louisiana. Association for Computational Linguistics.

\bibitem[{Wu et~al.(2023)Wu, Wang, Ye, and Kong}]{wu-etal-2023-self}
Zhiyong Wu, Yaoxiang Wang, Jiacheng Ye, and Lingpeng Kong. 2023.
\newblock \href {https://doi.org/10.18653/v1/2023.acl-long.79} {Self-adaptive
  in-context learning: An information compression perspective for in-context
  example selection and ordering}.
\newblock In \emph{Proceedings of the 61st Annual Meeting of the Association
  for Computational Linguistics (Volume 1: Long Papers)}, pages 1423--1436,
  Toronto, Canada. Association for Computational Linguistics.

\bibitem[{Yu et~al.(2022)Yu, Zhang, Xu, Zhang, Shen, and Zhang}]{yu2022cold}
Yue Yu, Rongzhi Zhang, Ran Xu, Jieyu Zhang, Jiaming Shen, and Chao Zhang. 2022.
\newblock Cold-start data selection for few-shot language model fine-tuning: A
  prompt-based uncertainty propagation approach.
\newblock \emph{arXiv preprint arXiv:2209.06995}.

\bibitem[{Yuan et~al.(2020)Yuan, Lin, and Boyd-Graber}]{cold}
Michelle Yuan, Hsuan-Tien Lin, and Jordan Boyd-Graber. 2020.
\newblock \href {https://doi.org/10.18653/v1/2020.emnlp-main.637} {Cold-start
  active learning through self-supervised language modeling}.
\newblock In \emph{Proceedings of the 2020 Conference on Empirical Methods in
  Natural Language Processing (EMNLP)}, pages 7935--7948, Online. Association
  for Computational Linguistics.

\bibitem[{Zhang et~al.(2015)Zhang, Zhao, and LeCun}]{agnews}
Xiang Zhang, Junbo Zhao, and Yann LeCun. 2015.
\newblock Character-level convolutional networks for text classification.
\newblock \emph{Advances in neural information processing systems}, 28.

\bibitem[{Zhang et~al.(2022)Zhang, Feng, and Tan}]{zhang-etal-2022-active}
Yiming Zhang, Shi Feng, and Chenhao Tan. 2022.
\newblock \href {https://aclanthology.org/2022.emnlp-main.622} {Active example
  selection for in-context learning}.
\newblock In \emph{Proceedings of the 2022 Conference on Empirical Methods in
  Natural Language Processing}, pages 9134--9148, Abu Dhabi, United Arab
  Emirates. Association for Computational Linguistics.

\bibitem[{Zhao et~al.(2021)Zhao, Wallace, Feng, Klein, and
  Singh}]{zhao2021calibrate}
Zihao Zhao, Eric Wallace, Shi Feng, Dan Klein, and Sameer Singh. 2021.
\newblock Calibrate before use: Improving few-shot performance of language
  models.
\newblock In \emph{International Conference on Machine Learning}, pages
  12697--12706. PMLR.

\end{thebibliography}
\bibliographystyle{acl_natbib}

\appendix

\section{Appendix}
\label{sec:appendix}
% \subsection{Relation with prompt bias}

\subsection{Implementation Details}
We use Pytorch and Huggingface Transformers in our implementation. We run all our evaluations on a single NVIDIA A40 GPU (48G). Our experiments should be also run on one single GPU of 24G. We access GPT-3 via the OpenAI API\footnote{https://openai.com/}. 

For experiments of GPT-2 XL in Table \ref{table:main}, we rerun the Random baseline due to differences in the training set mentioned in the repository\footnote{https://github.com/tonyzhaozh/few-shot-learning}. Nevertheless, our reimplemented results are similar to those reported in \cite{zhao2021calibrate}. Therefore, we report the reimplemented ones for a fair comparison with our proposed method.

\subsection{Dataset Details}
\label{app_dataset}
We show the statistics of datasets in Table \ref{table:app_data}. For SST-2~\cite{sst2}, AGNews~\cite{agnews}, TREC~\cite{trec}, and DBPedia~\cite{agnews}, we use their official test sets. For CB~\cite{cb}, RTE~\cite{rte}, MNLI~\cite{mnli}, SNLI~\cite{snli}, and BoolQ~\cite{boolq}, we use their original validation sets as test sets.
\begin{table}[htbp]
\begin{center}
\begin{tabular}{lllllccccccccccccccc}
\toprule[1pt]
\multicolumn{5}{l}{Dataset} & \multicolumn{5}{c}{\# Classes} & \multicolumn{5}{c}{\# Train} & \multicolumn{5}{c}{\# Eval}  \\ \hline
% gpt2 xl
\multicolumn{5}{l}{SST-2} & \multicolumn{5}{c}{2} & \multicolumn{5}{c}{6920} & \multicolumn{5}{c}{1821}  \\ 
% gptj 
\multicolumn{5}{l}{AGNews} & \multicolumn{5}{c}{4} & \multicolumn{5}{c}{120k} & \multicolumn{5}{c}{7.6k}  \\ 
\multicolumn{5}{l}{TREC} & \multicolumn{5}{c}{6} & \multicolumn{5}{c}{5452} & \multicolumn{5}{c}{500}  \\ 
\multicolumn{5}{l}{CB} & \multicolumn{5}{c}{3} & \multicolumn{5}{c}{250} & \multicolumn{5}{c}{56}  \\ 
\multicolumn{5}{l}{RTE} & \multicolumn{5}{c}{2} & \multicolumn{5}{c}{2490} & \multicolumn{5}{c}{277}  \\ 
\multicolumn{5}{l}{DBPedia} & \multicolumn{5}{c}{14}& \multicolumn{5}{c}{420k} & \multicolumn{5}{c}{70k}  \\ 
\multicolumn{5}{l}{MNLI} & \multicolumn{5}{c}{3}& \multicolumn{5}{c}{392k} & \multicolumn{5}{c}{9815} \\
\multicolumn{5}{l}{SNLI} & \multicolumn{5}{c}{3}& \multicolumn{5}{c}{549k} & \multicolumn{5}{c}{9842} \\
\multicolumn{5}{l}{BoolQ} & \multicolumn{5}{c}{2}& \multicolumn{5}{c}{9247} & \multicolumn{5}{c}{3270} \\

\bottomrule[1pt]
\end{tabular}
\caption{Statistics of evaluation datasets.}
\label{table:app_data}
\end{center}
\end{table}

\subsection{Template details for different tasks}
\label{app_template}
We show the templates used and corresponding label mappings for different tasks in Table \ref{tab:app_template}.

\subsection{Template Bias for different tasks across three LLMs}
\label{app_bias}
\begin{figure*}[htbp]
  \centering
  \subfigure[Template Bias of SST-2. Label Dictionary {1: Negative, 2: Positive}]{
      \includegraphics[scale=0.7]{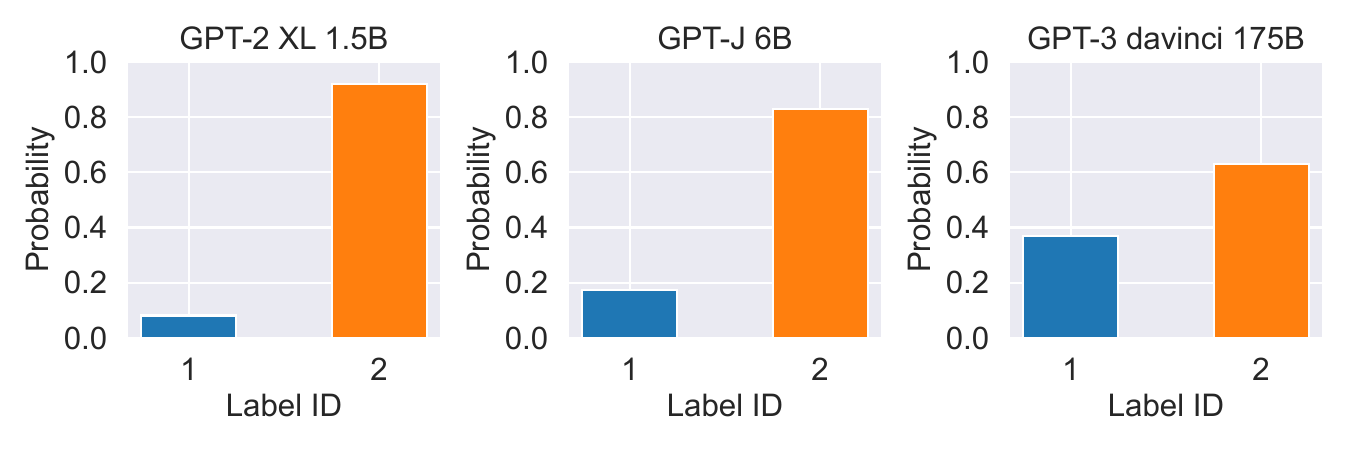}
  }
  \subfigure[Template Bias of AGNews. Label Dictionary {1: World, 2: Sports, 3: Business, 4:Technology }]{
      \includegraphics[scale=0.7]{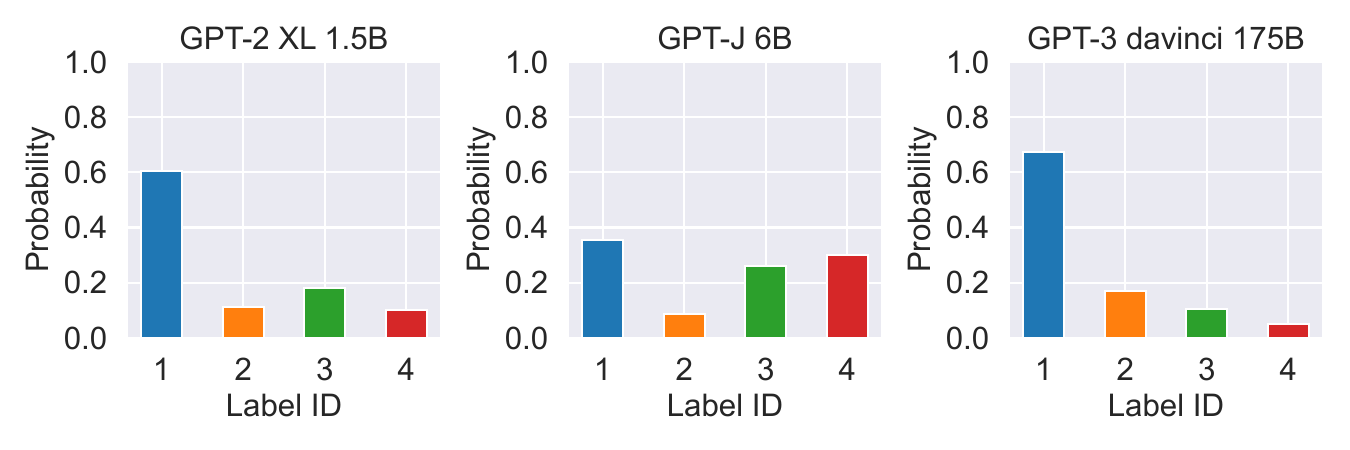}
  }
  \subfigure[Template Bias of TREC. Label Dictionary {1: Number, 2: Location, 3: Person, 4: Description, 5: Entity, 6:  Abbreviation}]{
      \includegraphics[scale=0.7]{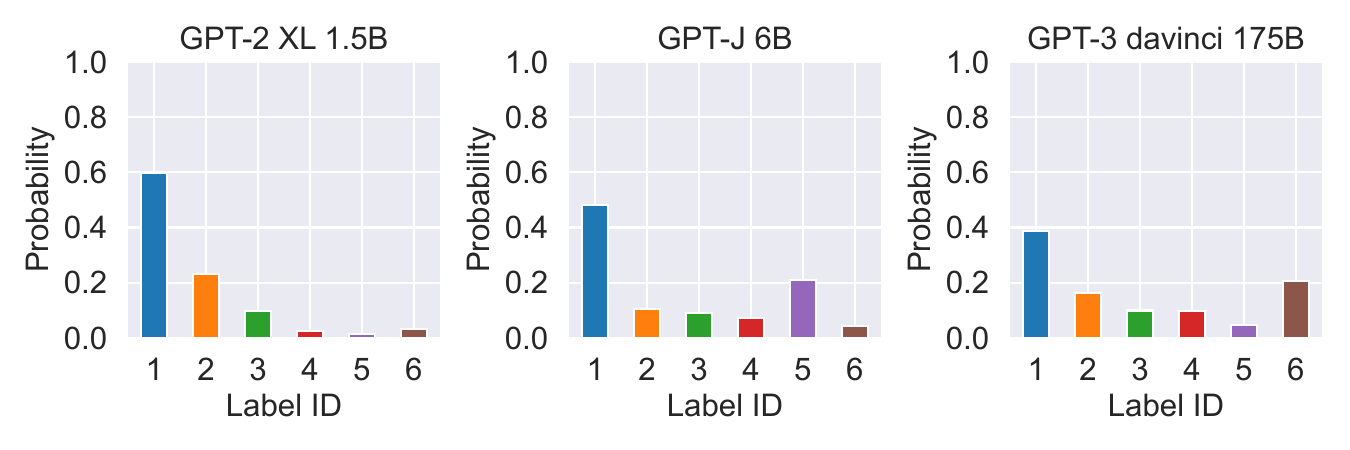}
  }

  \caption{ }
  \label{fig:bias1}

\end{figure*}

\begin{figure*}[htbp]
  \centering
  \subfigure[Template Bias of CB. Label Dictionary {1: True, 2: False, 3: Neither }]{
      \includegraphics[scale=0.7]{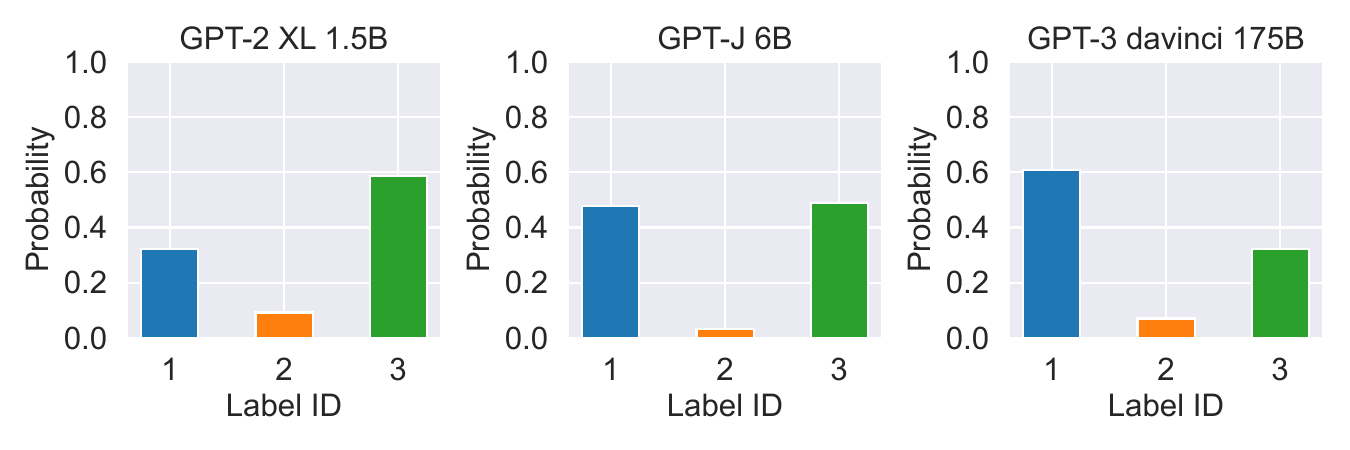}
  }
  \subfigure[Template Bias of RTE. Label Dictionary {1: True, 2: False}]{
      \includegraphics[scale=0.7]{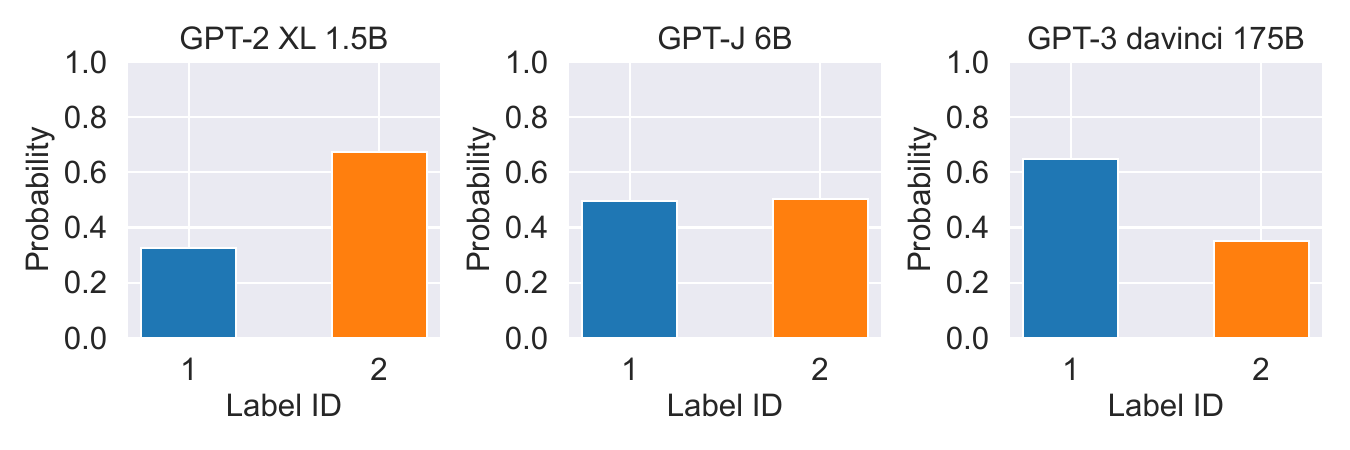}
  }
  \subfigure[Template Bias of DBPedia. Label Dictionary {1: Company, 2: School, 3: Artist, 4: Athlete, 5: Politician, 6: Transportation, 7: Building, 8: Nature, 9: Village, 10: Animal, 11: Plant, 12: Album, 13: Film, 14: Book }]{
      \includegraphics[scale=0.42]{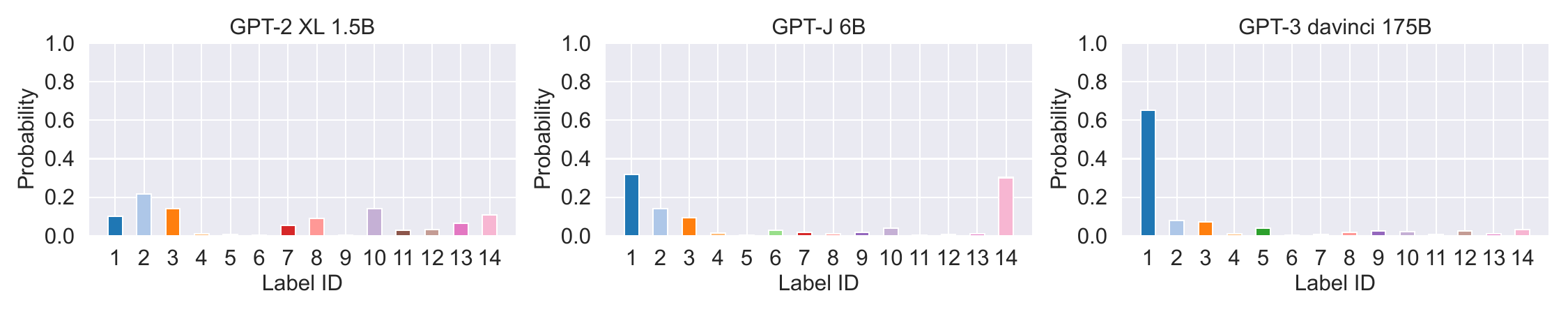}
  }
  \caption{ }
  \label{fig:bias2}

\end{figure*}

We plot the Template Bias of templates used for all tasks across three LLMs in Figure \ref{fig:bias1} and Figure \ref{fig:bias2}. It is observed that the template bias persists across different tasks and across LLMs with varying model sizes.  

\subsection{More experiments}
We consider broader NLI and commonsense reasoning tasks. Specifically, we conducted additional one-shot experiments on 3-way classification MNLI, 3-way classification SNLI, and 2-way classification BoolQ using GPT-J 6B. We evaluate them with the same setting in the main experiments. Table \ref{table:app_nli} shows the performance comparison between our method and baselines. It is observed that ICL without additional reasoning techniques performs poorly on MNLI, SNLI, and BoolQ, which aligns with prior work~\cite{chang2022careful,hongjin2023selective}. Nevertheless, our proposed method still outperforms other baselines on all three tasks.

\begin{table}[htbp]
\begin{center}
\begin{tabular}{llllcccccccccccc}
\toprule[1pt]
\multicolumn{4}{l}{Method} & \multicolumn{4}{c}{MNLI} & \multicolumn{4}{c}{SNLI} & \multicolumn{4}{c}{BoolQ} \\ \hline

\multicolumn{4}{l}{Random}  & \multicolumn{4}{c}{$35.9_{7.1}$} & \multicolumn{4}{c}{$33.7_{0.0}$} & \multicolumn{4}{c}{$58.3_{5.5}$} \\ 
\multicolumn{4}{l}{MaxEntropy}  & \multicolumn{4}{c}{$35.1_{5.2}$} & \multicolumn{4}{c}{$37.5_{1.3}$} & \multicolumn{4}{c}{$\textbf{61.1}_{2.7}$} \\ 
\multicolumn{4}{l}{MaxIG}  & \multicolumn{4}{c}{$36.1_{5.3}$} & \multicolumn{4}{c}{$37.1_{2.5}$} & \multicolumn{4}{c}{$56.6_{4.1}$} \\ 
\multicolumn{4}{l}{CBS MaxIG}  & \multicolumn{4}{c}{$\textbf{38.3}_{5.5}$} & \multicolumn{4}{c}{$\textbf{40.7}_{3.2}$} & \multicolumn{4}{c}{$\textbf{61.1}_{2.7}$} \\ 

\bottomrule[1pt]
\end{tabular}
\caption{One-shot performance on MNLI, SNLI, and BoolQ using GPT-J.}
\label{table:app_nli}
\end{center}
% \vspace{-5mm}
\end{table}

% \usepackage{graphicx}
\begin{table*}[]
\centering
\resizebox{\textwidth}{!}{%
\begin{tabular}{lll}
\hline
Dataset &
  Prompt Template &
  Label Mapping \\ \hline
SST-2 &
  \begin{tabular}[c]{@{}l@{}}Review: The movie is a desperate miscalculation.\\ Sentiment: Negative\\ \\ Review: I hate this movie.\\ Sentiment:\end{tabular} &
  Positive, Negative \\ \hline
AGNews &
  \begin{tabular}[c]{@{}l@{}}Article: Toronto Raptors Team Report - November 20,(Sports Network) - The Toronto\\ Raptors found themselves on the wrong end of a 101-94 decision against the red-hot \\ Seattle SuperSonics on Friday at the Air Canada Centre.\\ Answer: Sports\\  \\ Article: Pioneer replaces plasma TV power supplies, Certain Pioneer TVs have a faulty \\ power supply. An upgrade is available.\\ Answer:\end{tabular} &
  World, Sports, Business, Technology \\ \hline
TREC &
  \begin{tabular}[c]{@{}l@{}}Classify the questions based on whether their answer type is a Number, Location, Person,\\ Description, Entity, or Abbreviation.\\ \\ Question: Where is South Bend ?\\ Answer Type: Location\\ \\ Question: How many colors are there in the spectrum ?\\ Answer Type:\end{tabular} &
  \begin{tabular}[c]{@{}l@{}}Number, Location, Person, Description,\\ Entity, Abbreviation\end{tabular} \\ \hline
DBPedia &
  \begin{tabular}[c]{@{}l@{}}Classify the documents based on whether they are about a Company, School, Artist, Athlete, \\ Politician, Transportation, Building, Nature, Village, Animal, Plant, Album, Film, or Book.\\ \\ Article: Al Gamil is a privately held company based in Djibouti City Djibouti.\\ Answer: Company \\ \\ Article: Imperial Botanical Beach Hotel is a hotel in Entebbe Uganda.\\ Answer:\end{tabular} &
  \begin{tabular}[c]{@{}l@{}}Company, School, Artist, Athlete, Politician, \\ Transportation, Building, Nature, Village, \\ Animal, Plant, Album, Film, Book\end{tabular} \\ \hline
CB &
  \begin{tabular}[c]{@{}l@{}}Richard Breeden hadn't noticed that his new desk had just four telephone lines and one phone.\\ question: Richard Breeden's new desk had just four telephone lines and one phone. True, \\ False, or Neither?\\ answer: True\\ \\ "I know the one. Yes, it was good though I say it myself." But that doesn't mean I have to \\ be involved in this kind of nauseous business.\\ question: she has to be involved in this kind of nauseous business. True, False, or Neither?\\ answer:\end{tabular} &
  True, False, Neither \\ \hline
RTE &
  \begin{tabular}[c]{@{}l@{}}IKEA offers fantastic and affordable solutions for your home furnishing needs.\\ question: Ikea is a home. True or False?\\ answer: False\\ \\ I will take a brief vacation with some priest friends after Christmas and then I will go on \\ retreat at a monastery, Law, reading from a brief statement, told reporters.\\ question: Law said he plans to take a brief vacation after Christmas and later retreat to a \\ monastery. True or False?\\ answer:\end{tabular} &
  True, False \\ \hline
\end{tabular}%
}
\caption{Prompt template and label mapping for different tasks.}
\label{tab:app_template}
\end{table*}

\newpage

\end{document}